%% file: arxiv_iccv2025.tex
\documentclass[10pt,twocolumn,letterpaper]{article}

\usepackage[pagenumbers]{iccv} 


\definecolor{iccvblue}{rgb}{0.21,0.49,0.74}
\usepackage[pagebackref,breaklinks,colorlinks,allcolors=iccvblue]{hyperref}
\usepackage{multirow} 
\usepackage{makecell}
\usepackage{subcaption}
\usepackage[capitalize]{cleveref}
\usepackage {colortbl,array,xcolor}

\def\paperID{2381} 
\def\confName{ICCV}
\def\confYear{2025}
\newcommand{\ours}{CaptionSmiths\xspace}

\title{CaptionSmiths: Flexibly Controlling Language Pattern in Image Captioning}

\author{%
  Kuniaki Saito\thanks{OMRON SINIC X, kuniaki.saito@sinicx.com}      
   \and
   Donghyun Kim\thanks{Korea University}   
  \and
  Kwanyong Park\thanks{University of Seoul}
  \and
  Atsushi Hashimoto\textsuperscript{*}
  \and
  Yoshitaka Ushiku\textsuperscript{*}
}

\begin{document}
\maketitle

\input{chapters/00_abstract}
\input{chapters/01_intro}

\input{chapters/02_related}

\input{chapters/03_method}

\input{chapters/04_experiments}

\input{chapters/10_conclusion}

{
    \small
    \bibliographystyle{ieeenat_fullname}
    \bibliography{main}
}
\clearpage
\onecolumn
\appendix

\setcounter{figure}{0}
\setcounter{table}{0}
\setcounter{section}{0}
\renewcommand{\thefigure}{\Alph{figure}}
\renewcommand{\thetable}{\Alph{table}}
\renewcommand{\thesection}{\Alph{section}}

\input{appendix_content}

\end{document}

%% file: chapters/00_abstract.tex
\begin{abstract}
An image captioning model flexibly switching its language pattern, \eg, descriptiveness and length, should be useful since it can be applied to diverse applications. 
However, despite the dramatic improvement in generative vision-language models, fine-grained control over the properties of generated captions is not easy due to two reasons: (i) existing models are not given the properties as a condition during training and (ii) existing models cannot smoothly transition its language pattern from one state to the other. Given this challenge, we propose a new approach, CaptionSmiths, to acquire a single captioning model that can handle diverse language patterns. First, our approach quantifies three properties of each caption, length, descriptiveness, and uniqueness of a word, as continuous scalar values, without human annotation. Given the values, we represent the conditioning via interpolation between two endpoint vectors corresponding to the extreme states, e.g., one for a very short caption and one for a very long caption. Empirical results demonstrate that the resulting model can smoothly change the properties of the output captions and show higher lexical alignment than baselines. For instance, CaptionSmiths reduces the error in controlling caption length by 506\% despite better lexical alignment. Code will be available on \url{https://github.com/omron-sinicx/captionsmiths}.


\end{abstract}

%% file: chapters/01_intro.tex
\section{Introduction}
\label{sec:intro}
Image captioning is an important computer vision task, having a wide range of applications \eg, describing the surrounding environment to assist visually impaired individuals. We have seen dramatic improvements in this task~\cite{li2023blip,hu2022scaling} with the help of vision and language foundation models such as CLIP~\cite{radford2021learning} and Llama~\cite{touvron2023llama}. Recent trends in vision-language models focus on linking powerful image encoders with language models (LMs), leveraging large-scale image-text datasets~\cite{liu2024visual,zhu2023minigpt,liu2024improved,bai2023qwen,chen2025sharegpt4v}. 

\input{figs/teaser}
Tailoring a foundational model for a specific application can lead to a high-performance model, but it often requires adapting its output language patterns. For example, some users prefer detailed, lengthy captions, while others favor concise descriptions. These preferences are defined by human-interpretable criteria, such as caption length or the use of fine-grained versus general categories to describe a concept.
However, unlike \textit{wordsmiths}, humans skilled in crafting language, current captioning models struggle to control language patterns based on applications' demand.

Simply training powerful foundation models on diverse datasets does not offer precise control over these language patterns, as standard models tend to generate “average” captions from the training data, potentially losing the unique style of captions~\cite{wang2020diversity, luo2020analysis}. While training separate models for each application might mitigate this issue, it comes with significant costs, such as collecting image-caption pairs and training individual models for each application. To overcome these limitations, we aim to develop a single model capable of \textit{smoothly} adapting its behavior based on interpretable conditions as inputs.

Existing works tackle the challenge by conditioning the model on the pattern of the target caption~\cite{deng2020length, wang2023controllable, chen2022learning}, such as length and dataset type. Their models are trained to generate a caption given an image and the condition, such as clustering captions by their length and conditioning the model on the cluster index~\cite{deng2020length}. However, these works have three limitations: (i) their investigation is limited to a single image-captioning dataset, such as COCO, (ii) the conditioning is restricted to length alone, and (iii) it relies on a \textit{discrete} cluster index. The model with discrete conditioning still tends to generate an average caption within the specified cluster, neglecting the variation within the cluster. This means that the model can only transition between the center of one cluster and that of another, but struggles to represent the intermediate states between the two.

In this work, we propose an approach called \textit{\ours} to develop a single model capable of \textit{smoothly} adapting its behavior based on three properties: length, descriptiveness, and word uniqueness. This approach will enable caption generation with varying lengths, information densities, and rich vocabulary. While we build upon the concept of conditioned generation, the key challenges are (i) quantifying the conditions for each caption (beyond just length) and (ii) providing these quantified conditions to a language model for smooth control. To address these, we propose a novel method to automatically evaluate the three properties as scalar values for each caption, without requiring additional supervision. For example, descriptiveness is computed as the ratio of adjectives and nouns within a caption, normalized based on the overall statistics of the dataset.
Second, instead of using discrete conditioning, we employ the computed scalar value to linearly interpolate between two endpoint vectors corresponding to the extreme states, such as one for a very short caption (state A) and one for a very long caption (state B). As the scalar moves from 0 (fully state A) to 1 (fully state B), the condition vector smoothly transitions between these extremes. This continuous interpolation enables a gradual shift in model behavior, avoiding the abrupt jumps associated with discrete clusters. This parameterization avoids the need of hyper-parameter tuning in the number of clusters and is parameter-efficient. As shown in Fig.~\ref{fig:teaser}, our approach offers slider-like, flexible control over three properties. 


Our contributions are summarized as follows: \begin{enumerate} \item A novel framework for training an image captioning model that controls the language pattern of the output. \item Extensive experiments to confirm that our model can control the output pattern while maintaining semantic alignment with the image content. \item Empirical studies showing that the proposed conditioning achieves more precise control over the output pattern than existing approaches. \end{enumerate}

%% file: figs/teaser.tex
\begin{figure}[tp]
    \centering
    \includegraphics[width=\linewidth]{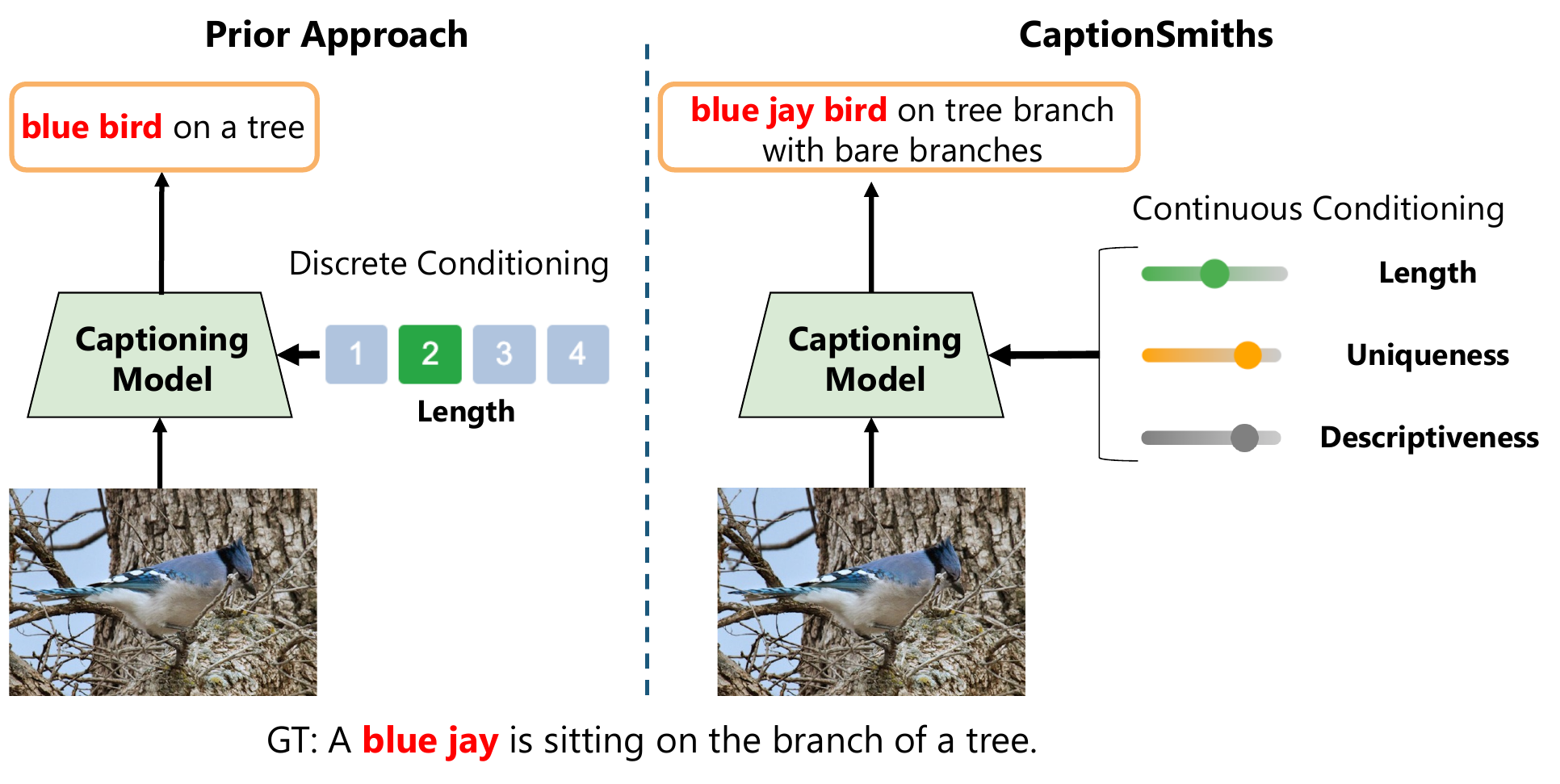}
    \vspace{-7mm}
    \caption{Left: Existing approach performs discrete state switching to control the caption length. Right: Our approach takes continuous values to control the length, uniqueness of words, and descriptiveness of the caption, achieving flexible and interpretable control for the generated caption.}
    \label{fig:teaser}
\end{figure}

%% file: chapters/02_related.tex
\section{Related Work}\label{sec:related}
\vspace{-2mm}
\textbf{Vision-language models.} 
Many vision-language foundation models have been introduced to address various vision tasks with a single model~\cite{liu2024visual,zhu2023minigpt,bai2023qwen,chen2025sharegpt4v,liu2024improved, Qwen-VL,Qwen2-VL}. These are typically pre-trained on a large dataset of image-caption pairs to train the connector module that maps image representations into token embeddings~\cite{liu2024visual}. They are then fine-tuned for various vision tasks, many of which require linguistic reasoning abilities. However, the flexibility to generate diverse captions using a single model has been overlooked.

\noindent\textbf{Controllable image captioning.}
Deng~\etal cluster captions by length and use the cluster index to control caption length~\cite{deng2020length}, while Dwibedi~\etal condition the model on length~\cite{dwibediflexcap}. However, both approaches are limited as they only control length and rely on discrete indices. Wang~\etal control caption style with a dataset-specific prompt~\cite{wang2023controllable}, but their method captures only the average caption, neglecting diversity. Chen~\etal use sentence embeddings to cluster captions and control generation with discrete indices, but their clusters lack interpretability~\cite{chen2022learning}. 
Kastner~\etal\cite{kastner2021imageability} control caption properties such as length and imageability, with the latter estimated using WordNet’s hierarchy, yet employ a discrete conditioning.
Kang~\etal condition the model based on the caption-image similarity for learning from noisy captions~\cite{kang2023noise}, also using discrete conditioning. Lu~\etal use a discrete reward token to condition the language model for specific tasks~\cite{lu2022quark}. In contrast, our method uses continuous scalar values to smoothly control caption generation, offering more precise control and generalizing across multiple captioning datasets. 
Chen~\etal condition a model with a scene graph to control caption output, which requires modules to obtain the scene graph from an image~\cite{chen2020say}. Although we acknowledge that there can exist important properties to control other than the three properties we choose, controlling the three can achieve an interesting application of image-captioning. 



\noindent\textbf{Model parameter merging.}
Model soup~\cite{wortsman2022model} linearly interpolates different models' parameters for better generalization. Prior work in the language field~\cite{vu2021spot,pfeiffer2020adapterfusion} employs this technique for better initialization or control model behavior. Our work is similar to these in that we employ the technique in the token parameter space. 
While the parameters in these works represent the properties of one \textit{task or dataset}, our work utilize the token parameter to let the model know the properties of \textit{an output caption}. 

%% file: chapters/03_method.tex
\input{figs/overall_pipeline}
\vspace{-2mm}
\section{Method}\label{sec:method}
\vspace{-2mm}

The overview of \ours is depicted in Fig.~\ref{fig:overall_pipeline}. We feed conditioning tokens that can control the property of the output captions during inference, which requires obtaining the conditioning tokens during training. Then, our training requires (i) computing the condition variables (\textit{Condition Calculator}) for each caption and (ii) embedding the variables into an LM's parameter space (\textit{Condition Encoding}). Our technical contributions lie in the two modules; \textit{Condition Calculator} enables to employ three properties as conditions while \textit{Condition Encoding} achieves slider-like control between states. Before explaining the two modules, we describe an overview of the baseline and our approach. 

\subsection{Overview}
\noindent\textbf{Model architecture.} As shown in Fig.~\ref{fig:overall_pipeline}, we employ the architecture of LLaVA~\cite{liu2024visual}, where the decoder-only language model ($f_{\phi}$) and vision encoder are connected with the projection model ($\mathbf{W}$). The image feature is extracted from the vision encoder, projected into the token embedding space by the projection, and fed into the language model with prompt embeddings. 

\noindent\textbf{Standard training.}
Given a pair of an image $I$ and a caption $c$ consisting of $T_{c}$ words, $w_{1}, w_{2}, ..., w_{T_{c}}$, the captioning models are trained with the following auto-regressive loss: 
\vspace{-5mm}
\begin{equation}
    \mathcal{L} = -\log p(c|I) = -\frac{1}{T_{c}} \sum_{t=0}^{T_{c}-1} \log p(w_{t+1} | w_{<t}, I). 
\end{equation}
The model learns to output an “average" caption which does not capture a language pattern specific to each caption~\cite{wang2020diversity}. To avoid the issue, we propose to prepend conditioning tokens to each caption for controllable generation. 

\noindent\textbf{Proposed Method.} 
Our approach has two core techniques. The first is preparing conditions for each caption. Given a set of captions, $C$, we compute a scalar value for each caption $c\in C$, which quantifies a certain caption's property and is normalized to lie within the range [0,1]. The second is encoding the scalar into the LM's token embedding space by linearly interpolating two learnable parameters. Given the conditioning embeddings, our approach minimizes the following loss: 
\vspace{-3mm}
\begin{equation}
    \mathcal{L} = -\log p(c|I) = -\frac{1}{T_{c}}\sum_{t=0}^{T_{c}-1} \log p(w_{t+1} | w_{<t}, I, f(c)), 
\end{equation}
where $f(c)$ denotes the function that converts the caption into learnable tokens. 

\input{figs/condition_compute}
\subsection{Condition Calculator}\label{sec:condition_prep}
We aim to control three key properties of captions: \textit{length}, \textit{descriptiveness}, and \textit{uniqueness}. The optimal \textit{length} of a caption varies across applications—some scenarios favor long descriptions, while others require concise ones. \textit{Descriptiveness} also depends on the application; a caption can provide a highly detailed depiction of a scene or a more minimalistic summary. For \textit{uniqueness}, the choice between general and domain-specific terminology is often user-dependent. For instance, children may struggle to understand specialized terms. At the same time, we recognize that other quantifiable properties, such as alignment with image content~\cite{kang2023noise} or structural coherence~\cite{chen2020say}, could also be valuable to control. Expanding our approach to incorporate additional properties remains an avenue for future work.
The workflow illustrated in Fig.~\ref{fig:condition_compute} consists of three steps: (i) property assessment, which calculates numerical values for each caption; (ii) decorrelation of these values; and (iii) normalization of the values to the range [0, 1] for ease of Condition Encoding (Sec.\ref{sec:condition_encoding}).

\noindent\textbf{Step 1: Property assessment.}

\begin{itemize}
    \item \noindent\textbf{Length.}
The expected length of the caption should differ by application and controlling the length can be critical. We tokenize each caption using a pre-defined tokenizer\footnote{We employ LLama's tokenizer since we mainly employ LLaVa for experiments} and utilize the number of tokens as a conditioning, represented as $L_{c}$.  
\item \noindent\textbf{Descriptiveness.} 
We define \textit{descriptiveness} as the ratio of nouns and adjectives in a caption. Intuitively, this measures the density of the information in a caption. Note that captions with the same length do not necessarily have the same descriptiveness. For instance, given two captions describing a dog running in a park: “I can see a dog." and “Dog in a park", the latter conveys richer information, \ie, “dog" and “park". We then compute this metric for a caption $c$ as follows: 
\begin{equation}
D_c = \frac{1}{T_c}\sum_{t=1}^{T_c} I[w_t \in \text{ADJ} \cup \text{NOUN} \setminus \mathcal{V}_{\text{excl}}],
\end{equation}
where $T_{c}$ and $w_t$ denote the number of words in a sentence and $t'$th word in a sentence respectively, and $I[w_t \in \text{ADJ} \cup \text{NOUN}\setminus \mathcal{V}_{\text{excl}}]$ is an indicator function that checks whether a word is an adjective or noun except those in the excluded set $\mathcal{V}_{\text{excl}}$\footnote{Spacy~\cite{spacy2} is used to detect adjectives and nouns.}. The excluded set includes non-descriptive nouns such as “image" and “photo".
\item \noindent\textbf{Uniqueness of vocabularies.} 
The vocabulary used to describe an entity can vary depending on the individual. For example, a golden retriever may be referred to simply as a “dog" or more specifically as a “golden retriever". Whether to use a general or fine-grained term depends on the context and application. Therefore, we aim to control this variability. One way to quantify this variation is by leveraging a hierarchical structure, such as WordNet~\cite{miller1995wordnet}, which organizes entities into predefined categories. However, such databases may not capture all the vocabulary present in image captions. Motivated by the observation that captions tend to favor general terms over more specific ones, we introduce the concept of vocabulary uniqueness as a condition. Specifically, given a set of captions in a dataset, $C$, we first create a dictionary where each key is a word, and its corresponding value is the word's frequency in $C$. The vocabulary uniqueness metric for a caption $c$ is then computed as follows: 
\begin{equation} 
U_c = \frac{1}{T_c} \sum_{t=1}^{T_c} \frac{1}{F(w_t)}, 
\end{equation} where $F(w_t)$ denotes the frequency of the word $w_t$ in $C$. $U_c$ increases as the caption includes more rare words.
\end{itemize}

\noindent\textbf{Step 2: Decorrelation (optional).} 
Decorrelating the computed values can be critical to controlling properties independently. Fortunately, the correlation between the values computed above is small\footnote{The biggest correlation is -0.11.}.
However, when two values are highly correlated, adjusting one can unintentionally affect the other. Removing the correlation while maintaining interpretability is a challenging task. Then, we provide a solution as an optional design. We first set the caption length as the primary control target and leave it unchanged. Next, we decorrelate uniqueness based on the length, followed by further decorrelation of density using both the length and the processed uniqueness. 
We use linear regression to achieve this process~\cite{zou2005regularization}, where we train a regression model to predict property A from property B and subtract the predicted value from A. The details of this procedure and ablation study on the decorrelation are available in the appendix.



\noindent\textbf{Step 3: Normalization.} 
For convenience in Sec.~\ref{sec:condition_encoding}, we normalize the values computed above into the range between 0 to 1. Although there are several design choices, we simply normalize the values by the maximum value, \eg, $\bar{D}_c$: $\frac{D_c}{max_{c} D_c}$, can convert all values into [0, 1], which maintains the relative interval between different values. 
\input{tables/exp_diverse_datasets}

\subsection{Condition Encoding}\label{sec:condition_encoding}
Previous approaches~\cite{deng2020length, kang2023noise} condition LMs with a discrete index to obtain an embedding, where each group corresponds to a learnable token. However, such models can only jump between states and struggle with intermediate ones. To address this, using many groups can represent multiple states, but increases parameters and requires tuning the number of groups. Instead, we use a computed scalar value to linearly interpolate between two endpoint learnable parameters, which correspond to the extreme states—\eg, one for a very short caption (state A) and one for a very long caption (state B). As the scalar moves from 0 (fully state A) to 1 (fully state B), the condition vector transitions smoothly between these extremes. For instance, we compute the length conditioning embedding for $c$, $E_{c}^{L}$, as follows: \begin{equation}~\label{eq:encoding} E_{c}^{L} = \bar{L}_c * E_{1}^{L} + (1 - \bar{L}_{c}) * E_{0}^{L}, \end{equation} where $E_{0}^{L}$ and $E_{1}^{L}$ are two endpoint vectors, each with $d$ dimensions, representing the states of \textit{very short} and \textit{very long}, respectively. This formulation offers advantages over discrete conditioning in both parameter and sample efficiency. Our approach increases the number of parameters by only $2 \times d$, whereas discrete conditioning increases the number of parameters by $k \times d$ ($k \geq 2$), with $k$ needing to be large for flexible control. Additionally, our method allows for the training of the two endpoint parameters on nearly all samples, while the discrete approach trains conditioning parameters based on samples from a single group. Results in Sec.~\ref{sec:analysis} suggest that increasing the number of groups in a discrete approach improves the control over output captions but reduces lexical alignment, likely due to inefficiencies in training examples. In contrast, our approach avoids such a trade-off.

We finally minimize the objective below during training: 
\begin{equation}
    \mathcal{L}_{c} = -\log p(c|I) = \sum_{t=1}^{T} p(w_{t+1} | w_{<t}, I, E_{c}^{L}, E_{c}^{D}, E_{c}^{U}). 
\end{equation}

\noindent\textbf{Connection with MLP-based condition encoding.} 
An alternative approach involves using a multi-layer perceptron (MLP) that takes a scalar as input. The simplest form of this method, i.e., using a single linear layer, is equivalent to our approach. Let the weight and bias in the linear layer be $w \in \mathbb{R}^{d}$ and $b \in \mathbb{R}^{d}$. Given a condition $L_c$, the operation is expressed as: 
\begin{equation}
   E_{c}^{L} =  L_c * w + b = L_c * (w+b) + (1-L_{c}) * b
\end{equation}

This equation is equivalent to Eq.~\ref{eq:encoding}, under the conditions that $w+b = E_{1}^{L}$,  $b=E_{0}^{L}$. Since we normalize the condition to the range [0, 1], this operation can be interpreted as linear interpolation between two points, $E_{1}^{L}$ and $b=E_{0}^{L}$ (model merging~\cite{wortsman2022model}), offering a clearer understanding of the roles of these two vectors. Additionally, one of our contributions is the empirical evaluation of the advantages of continuous conditioning over discrete conditioning.

\subsection{Inference}\label{sec:inference}
As shown in the right of Fig.~\ref{fig:overall_pipeline}, the user can directly feed conditioning values or feed an example-style sentence as a reference during inference. In a real application, for the length and descriptiveness, the user can feed the desired values, \eg, length: 10, and the corresponding condition calculator can convert the value into a conditioning scalar. Specifying the uniqueness value is not very intuitive since we employ the stats computed in the dataset. However, this issue can be addressed by displaying some example sentences and corresponding uniqueness values. 

%% file: figs/overall_pipeline.tex
\begin{figure*}[t]
    \centering
    \includegraphics[width=0.9\linewidth]{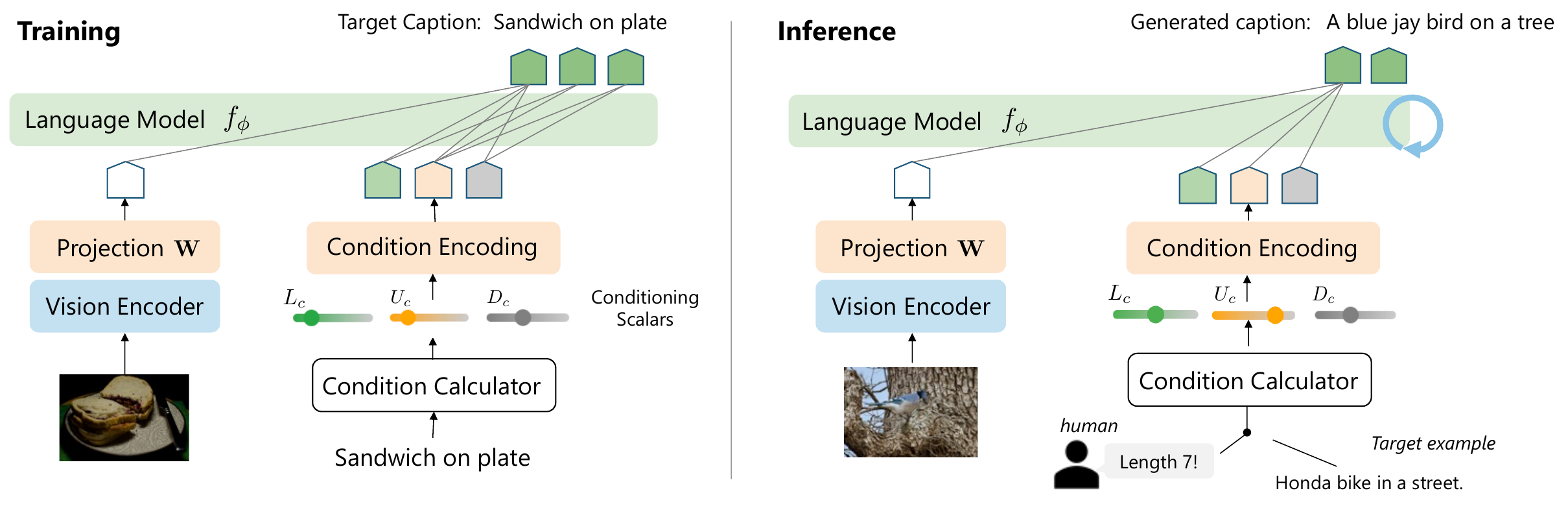}
    \vspace{-3mm}
    \caption{Overview during training (left) and inference (right). \textbf{Left: } We compute conditioning values for each caption in \textit{Condition Calculator} (Sec.~\ref{sec:condition_prep}). The conditions are converted into token embeddings via \textit{Condition Encoding} (Sec.~\ref{sec:condition_encoding}). Then, the learnable parameters are trained to control the language pattern in the output caption. {\textbf{Right}: } In inference, users can either manually specify the conditioning scalars or employ an example language pattern as a sentence. }
    \label{fig:overall_pipeline}
\end{figure*}

%% file: figs/condition_compute.tex
\begin{figure}[tp]
    \centering
    \includegraphics[width=\linewidth]{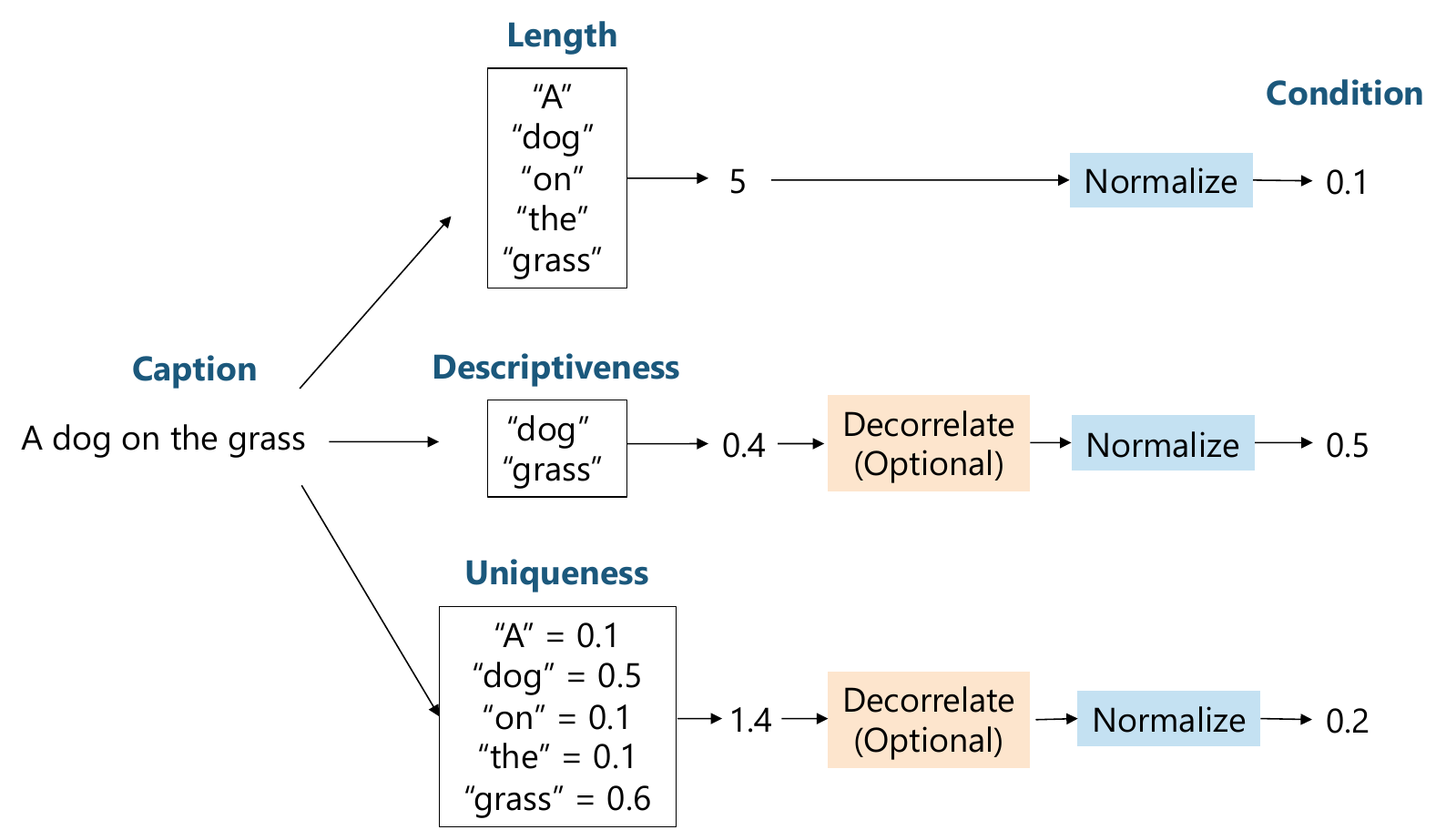}
    \vspace{-5mm}
    \caption{\textit{Condition Calculator} (Sec.~\ref{sec:condition_prep}) consists of three stages: (i) property assessment computes quantifies each caption's three properties, (ii) descriptiveness and uniqueness values are decorrelated, and (iii) values are normalized by the stats of all captions.}
    \label{fig:condition_compute}
\end{figure}

%% file: tables/exp_diverse_datasets.tex
\begin{table*}[t!]
    \begin{center}
    \scalebox{0.8}{
    \begin{tabular}{l|c|c|cccc|cccc|cccc}
        \toprule
        \multirow{2}{*}{Models}  &\multirow{2}{*}{\makecell{Conditioning \\(Prompting)}} &\multirow{2}{*}{\makecell{Total \\Params{$^\dagger$}}}&\multicolumn{4}{c|}{\multirow{1}{*}{MSCOCO (Short)}} &\multicolumn{4}{c|}{\multirow{1}{*}{LN COCO (Middle)}}&\multicolumn{4}{c}{\multirow{1}{*}{Docci (Long)}} \\           
        &&&B@4 &M &C &R &B@4 &M &C &R &B@4 &M &C &R \\\hline              \rowcolor[gray]{0.85}\multicolumn{15}{l}{\small 
       \textit{\textbf{Open-source models}}} \\ \hline
       LLaVA-1.5 & Vanilla &7.1B&1.0&27.1&0.0&12.6&1.4&26.2&1.1&16.8&2.0&16.9&2.4&17.6\\ 
       Blip-3~\cite{xue2024xgen} &  
       Concise/Detailed&4.6B&8.2&28.5&57.5&36.4&0.4&8.3&2.0&15.0&4.7&26.5&3.6&22.6\\   
              Qwen2-VL-7B~\cite{Qwen2-VL} & Concise/Detailed &8.3B& 9.9&34.4&84.3&39.7&0.4&10.3&3.8&17.6&5.4&28.3&5.5&23.0\\\hline
   
       \rowcolor[gray]{0.85}\multicolumn{15}{l}{\small 
       \textit{\textbf{Specialized models}}} \\ \hline
       Vanilla Supervised& N / A &21.3B&11.6&36.6&96.6&39.9 &9.6&33.7&23.6&32.3&4.7&22.8&1.4&22.4\\
       Fine-tuned LLaVA-1.5& N / A &21.3B&11.8&36.9&98.3&40.1&9.6&33.9&23.9&32.7&8.2&28.7&9.1&27.1\\
       \hline
        \rowcolor[gray]{0.85}\multicolumn{15}{l}{\small 
        \textit{\textbf{Trained on mixed dataset}}} \\ \hline
        Vanilla &Vanilla&7.1B&1.1&24.2&13.2&18.1&8.6&31.4&21.7&30.5&6.4&25.4&7.5&24.6\\
        Concap~\cite{wang2023controllable}&Dataset-aware&7.1B&\textbf{11.5}&36.7&95.9&\textbf{39.9}&\textbf{9.6}&33.5&23.5&32.3&7.2&26.8&8.3&26.0\\
        \ours &Target-aware&7.1B&11.4&\textbf{38.8}&\textbf{104.8}&39.8&\textbf{9.6}&\textbf{36.9}&\textbf{37.4}&\textbf{32.8}&\textbf{9.1}&\textbf{32.2}&\textbf{29.7}&\textbf{26.9}\\
        \bottomrule
    \end{tabular}
    }
    \end{center}
    \vspace{-0.55cm}
    \caption{
        \small Caption generation performance comparison with baselines on MSCOCO and nocaps datasets.
        B@4, M, C, and R mean BLEU@4, METEOR, CIDEr, and Rouge-L metrics, respectively.
        Numbers in \textbf{bold} indicate the best method. $^\dagger$ show the total number of parameters used to assess the three datasets (Specialized model needs a 7.1B model for each dataset).
    }
    \vspace{-0.3cm}
    \label{table:exp_diverse}
\end{table*}

%% file: chapters/04_experiments.tex
\section{Experiments}
We evaluate captions generated by our approach from two perspectives: (i) flexible control over three properties and (ii) lexical alignment with ground-truth sentences. We do not aim for a state-of-the-art captioning model for each dataset. Details are provided in the appendix, where tables and figures are numbered alphabetically. \ours uses minimal modifications to LLaVA~\cite{liu2024visual} for training. We will publish our code used for experiments upon acceptance.

\subsection{Setup}

\noindent\textbf{Model.}
We employ CLIP~\cite{radford2021learning}(ViT-L) as the image encoder and LLama-2 7B~\cite{touvron2023llama} as the decoder following LLaVA~\cite{liu2024visual}.


\noindent\textbf{Evaluation protocol.} 
We use a ``target-caption aware'' evaluation~\cite{chen2022learning, deng2020length}, where the model is conditioned on the desired caption properties, such as length, descriptiveness, and uniqueness. We compute the values of these properties using ground-truth captions and feed them to our model. Then, we assess the alignment between the target and generated captions. In the standard evaluation, one caption is matched to multiple ground-truth captions, with the best match being used as the alignment score. In contrast, our ``target-caption aware'' evaluation applies a one-to-one assessment to ensure performance aligns with the condition, using BLEU~\cite{papineni2002bleu}, Meteor~\cite{banerjee2005meteor}, CIDER~\cite{vedantam2015cider}, and Rouge~\cite{lin2004rouge} to measure lexical alignment.

\noindent\textbf{Datasets.}
We assess \ours when trained on image captioning datasets with rich length, detail, and vocabulary diversity. For this, we select several datasets, as shown in Table \textcolor{iccvblue}{A}, including LN COCO~\cite{pont2020connecting}, Detail23K~\cite{liu2024visual}, Docci~\cite{onoe2025docci}, Laion-COCO~\cite{laion_coco}, COCO~\cite{lin2014microsoft}, and Monkey~\cite{li2024monkey}, totaling 1.3M image-caption pairs.

\noindent\textbf{Baselines.}
\begin{itemize} 
\item Vanilla: We train LLaVA model using the same data without conditioning to see the effect of conditioning. 
\item Concap~\cite{wang2023controllable}: We follow~\cite{wang2023controllable} by using dataset-specific input prompts, replacing language prompts with dataset-specific token parameters. This baseline applies dataset-specific conditioning at test time. 
\item Specialized models: We train models on each evaluation dataset to compare against specialized captioning models. LLaVA-1.5~\cite{liu2024visual} is fine-tuned on each dataset, and a vanilla model is trained on each individually. Outperforming these models is not expected, as our model is designed to handle diverse datasets with a single model. 
\item Open-source models: Some open-source models, like Blip-3~\cite{xue2024xgen} and Qwen-2-VL~\cite{Qwen-VL}, accept caption style specifications. We use prompting to control these models and generate \textit{concise/detailed} captions. For comparison, we also include results from LLaVA-1.5~\cite{liu2024visual}. \end{itemize}

\subsection{Results}
\noindent\textbf{\ours exhibits high lexical alignment in diverse datasets.} 
We evaluate the lexical alignment between generated and ground-truth captions. Since the used metrics penalize caption length, this evaluation considers both word-level alignment and caption length. Table~\ref{table:exp_diverse} shows performance on three datasets to assess diverse caption lengths. \ours outperforms \textit{Vanilla}, trained on the same data without conditioning on target captions, highlighting the importance of conditioning. \textit{Vanilla} performs poorly on COCO due to its tendency to generate long captions. While \textit{Concap}~\cite{wang2023controllable} is comparable to \ours on COCO, it significantly underperforms on others, demonstrating that \ours better unifies knowledge across datasets. \ours is comparable to or even outperforms \textit{specialized models} tailored to each dataset. A comparison to open-source VLLM further supports \ours' superiority. These results indicate that \ours generalizes well across datasets with different caption styles.

\input{figs/length_vary_mainexp}
\input{figs/descriptive_clipscore}

\noindent\textbf{Ours can cover a diverse range of lengths.} 
Figure~\ref{fig:vary_length} shows the effect of varying the length conditioning value while keeping the other two conditions fixed. The results demonstrate that \ours covers diverse lengths while the baselines' coverage is limited. Additionally, we observe that the standard deviation is higher for longer captions, probably due to the limit of long captions in training data, which highlights the importance of dataset diversity.

\noindent\textbf{Increasing descriptiveness enriches caption information.} 
We study whether increasing the descriptiveness value enhances the richness of the generated descriptions. We expect that raising this value will produce captions with more descriptive words that are semantically aligned with the image. We use ClipScore~\cite{hessel2021clipscore}, which measures the alignment between the image and its caption using a CLIP model, on the COCO dataset. Figure~\ref{fig:clip_score} varies the descriptiveness conditioning while keeping the other two values fixed during caption generation. The results show that increasing the value improves the ClipScore, confirming that descriptiveness controls the richness of the caption's information as anticipated. Interestingly, captions generated with high descriptiveness conditions outperform the ground-truth captions. Figure~\ref{fig:example_descriptive} illustrates two examples with varying descriptiveness scores. As the descriptiveness increases, the generated captions tend to include more objects or adjectives, providing a more detailed explanation of the image.

\input{figs/example_descriptive}

\input{figs/uniqueness_evaluation}

\input{figs/example_uniqueness}

\noindent\textbf{Increasing uniqueness covers fine-grained category words.} 
We evaluate the impact of varying uniqueness conditioning, which controls the vocabulary used to describe a concept by choosing either a general or more fine-grained term. To assess this, we use fine-grained classification datasets: CUB~\cite{WahCUB_200_2011}, Stanford Dogs~\cite{khosla2011novel}, and Stanford Cars~\cite{krause20133d}. We compute the recall for fine-grained categories based on whether the generated caption includes the category name for each image. Figure~\ref{fig:uniqueness} shows the results of increasing uniqueness conditioning while keeping the other values fixed. The results demonstrate that higher uniqueness improves recall by generating more fine-grained descriptions. Figure~\ref{fig:example_uniqueness} visualizes the generated captions. As the uniqueness conditioning increases, the captions include more specific terms, such as \textit{blue jay} and \textit{cyclist}.

\input{tables/self_retrieval}
\input{figs/vocabulary_bar}

\noindent\textbf{Generated captions are uniquely connected to a source image.} 
We perform a self-retrieval evaluation following~\cite{kang2023noise} and assess whether the generated caption can correctly retrieve the input images from COCO, using a CLIP model. Table~\ref{table:self_retrieval} shows that \ours outperforms the baselines, especially surpassing Concap, which is trained on the same data without sample-specific conditioning. This suggests that our generated captions better capture image-specific features. In contrast, the baselines fail to account for the unique properties of captions during training, leading to more generic captions that resemble the average training data, making self-retrieval more challenging.

\noindent\textbf{\ours has rich vocabularies.} 
We further analyze the ratio of unique words in the generated captions using the COCO validation split, specifically computing the ratio of unique words to the total number of generated words. Figure~\ref{fig:vocabulary_var} shows that \ours generates a significantly more diverse vocabulary than the baselines. In contrast, captions from the baseline methods exhibit limited vocabulary diversity, which may hinder self-retrieval performance. This issue likely stems from the standard training process, which does not account for the specific characteristics of the target caption. These findings, along with the results in Table~\ref{table:self_retrieval}, confirm that captions generated by \ours more effectively describe images and better align with their content. Figure~\ref{fig:x_uniq_y_recall_vocab} illustrates the relationship between unique words and recall in self-retrieval as the uniqueness condition increases. These results suggest that increasing the uniqueness condition both diversifies the vocabulary and enhances alignment with the image content.

\input{figs/x_uniq_y_recall_vocab}

\noindent\textbf{Three conditions are well-disentangled.} 
In our framework, we expect the model to maintain the other two properties when varying one condition and fixing the others. Figure~\textcolor{iccvblue}{A} studies how varying one condition affects other properties, showing that the change in caption length when varying uniqueness conditions, with a change of just 1-2 tokens, which is minimal. In the other two cases, where we visualize the range of properties computed from COCO validation captions as a reference, the changes are also negligible. Although not perfect, our conditioning achieves well-disentangled control over the three properties.



\subsection{Analysis}~\label{sec:analysis}
In this section, we aim to see the effectiveness of the proposed approach in one dataset, COCO~\cite{lin2014microsoft}. Unlike the experiments on a mixture of captioning datasets above, experiments on one dataset ease the analysis of diverse aspects. We employ a COCO training split to train all models and evaluate models on the validation split. 
\input{tables/coco_only}

\noindent\textbf{Comparison with length-control baselines.} 
Table~\ref{tab:coco_only} compares \ours with the length-controlling baseline, LC~\cite{deng2020length}. To ensure a fair comparison, we modify the baseline to use the same vision encoder and language decoder as ours. The modified approach groups captions into five length-based categories and applies conditional training. The results show that \ours outperforms the baseline, indicating that controlling the descriptiveness and uniqueness of vocabulary improves alignment with the ground-truth captions.
\input{tables/length_difference}

\noindent\textbf{Continuous parameterization vs. discrete one.} 
Similar to the length-control baseline, we apply discrete grouping to the three properties and compare it with our continuous parameterization. Table~\ref{tab:length_difference} presents the results for different numbers of groups, showing that the continuous approach achieves better alignment with conditioning and lexical alignment with the ground-truth captions. Increasing the number of groups improves conditioning alignment, as evidenced by the lower length mismatch, but at the cost of decreased alignment with the ground-truth. This performance degradation can be attributed to the reduction in the number of samples used to train each learnable token, as discussed in Sec.~\ref{sec:condition_encoding}. 
Figure~\textcolor{iccvblue}{D} shows the alignment with ground-truth captions in terms of length. Captions generated by \ours align better than those generated with discrete conditioning, which groups captions of different lengths into the same category, preventing the model from capturing their unique lengths during generation.

\noindent\textbf{List of the empirical results in the appendix.}
Below are the summaries of results not mentioned in the main draft.
\begin{itemize}
    \item Hallucination in long-captions: As discussed in previous study~\cite{hirota2024descriptive}, the model suffers from hallucination in generating long-captions, which is a limitation of \ours.
    \item Zero-shot evaluation: Evaluation using Nocaps~\cite{agrawal2019nocaps} and Vizwiz~\cite{gurari2020captioning} shows that \ours outperforms other open-source models.
    \item Evaluation with ChatGPT: We show that our criterion of descriptiveness is aligned with ChatGPT's criterion. 
    \item Ablation study to control three properties.
\end{itemize}

%% file: figs/length_vary_mainexp.tex
\begin{figure}[t]
    \centering
    \includegraphics[width=0.85\linewidth]{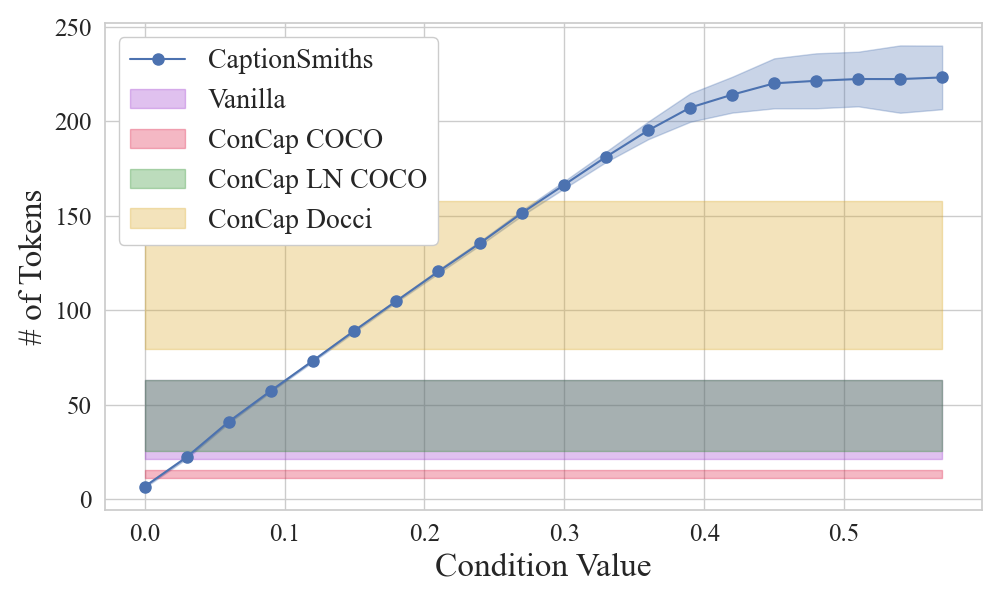}
    \vspace{-3mm}
    \caption{Results of varying length conditioning. The Y-axis indicates the number of generated tokens for each conditioning value. The range of length covered by each baseline (mean $\pm$ standard deviation) is highlighted by different colors.}
    \label{fig:vary_length}
\end{figure}

%% file: figs/descriptive_clipscore.tex
\begin{figure}[t]
    \centering
    \includegraphics[width=0.7\linewidth]{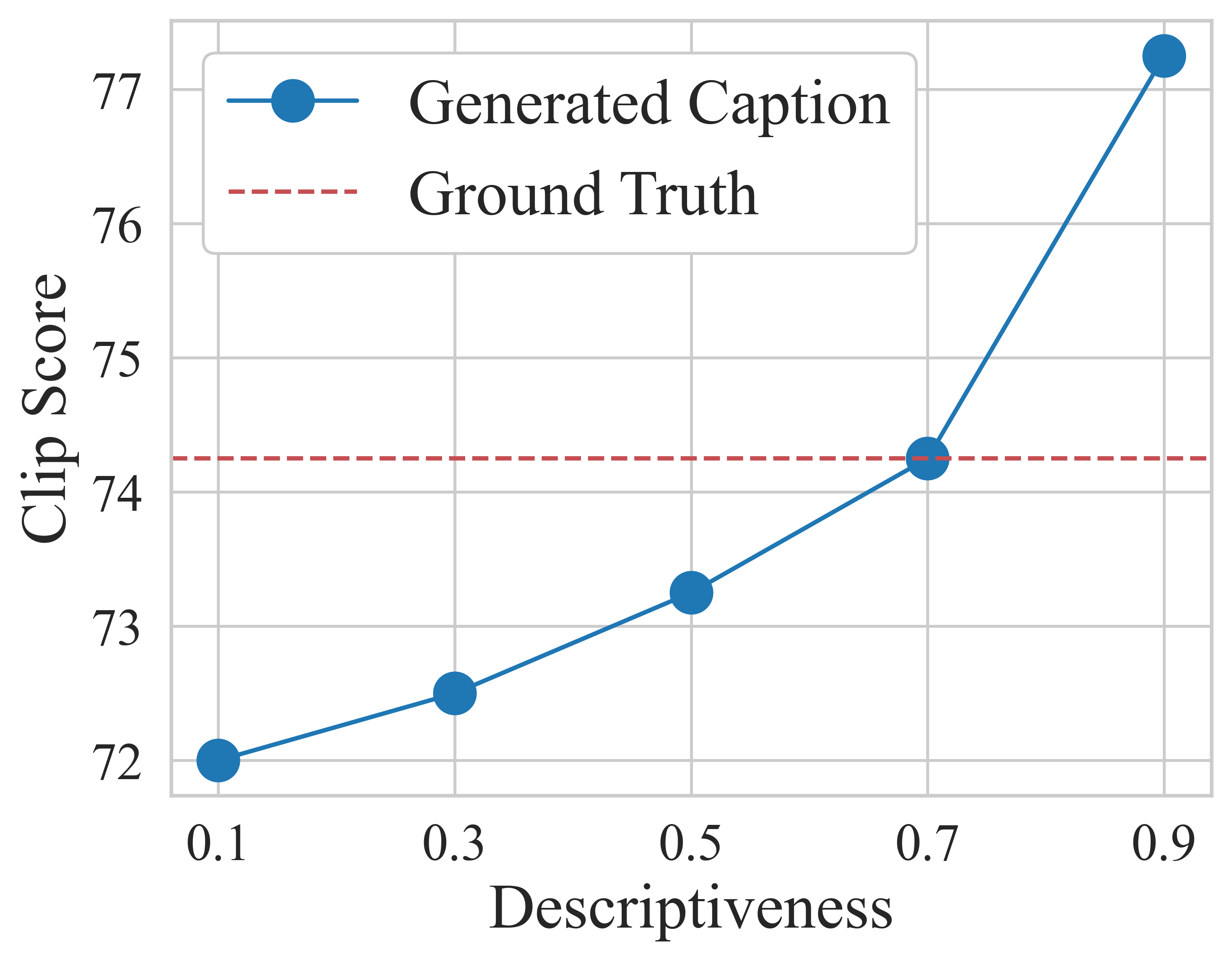}
    \vspace{-3mm}
    \caption{Increasing descriptiveness improves clip score, meaning that our descriptiveness enriches information in the caption and it is aligned with image content.}
    \label{fig:clip_score}
\end{figure}

%% file: figs/example_descriptive.tex
\begin{figure*}[htbp]
    \centering
    \includegraphics[width=\linewidth]{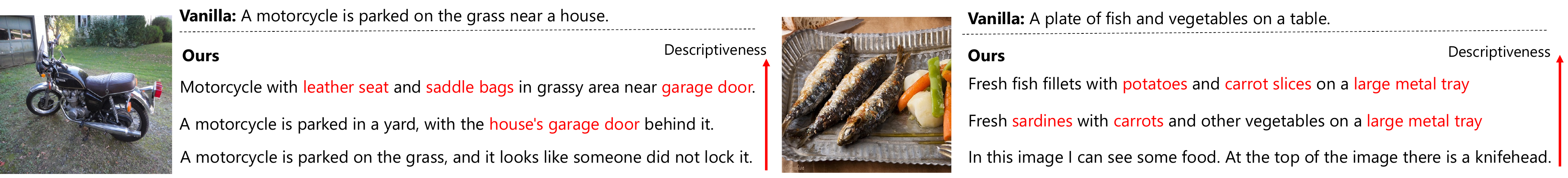}
        \vspace{-7mm}
    \caption{Generated captions by increasing descriptiveness conditioning value while fixing the other two. More objects are described.}
    \label{fig:example_descriptive}
\end{figure*}

%% file: figs/uniqueness_evaluation.tex
\begin{figure*}[htbp]
\vspace{-1mm}
    \centering
    \begin{subfigure}[b]{0.3\textwidth} 
        \centering
        \includegraphics[width=0.9\textwidth]{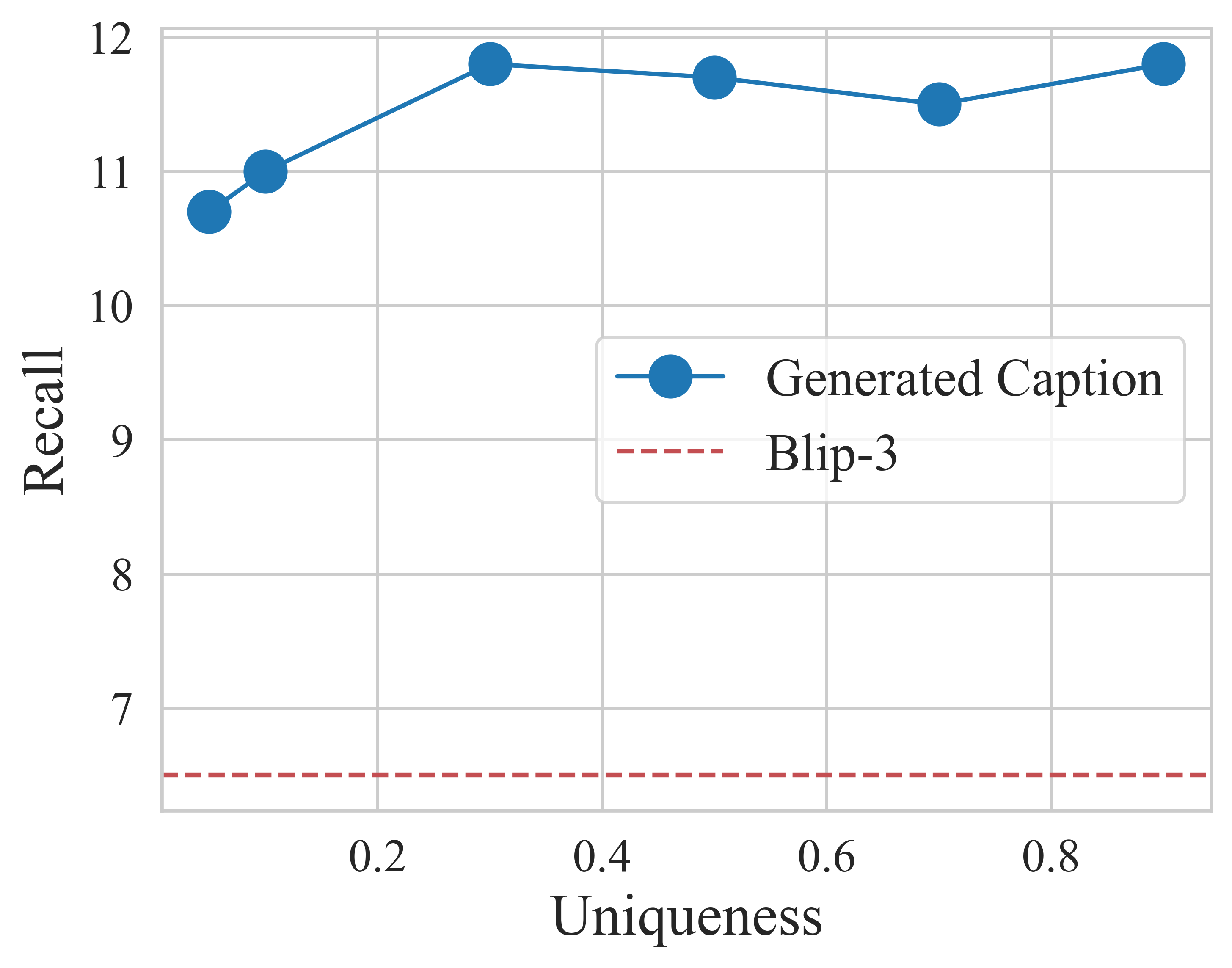} 
        \caption{CUB}
        \label{fig:sub1}
    \end{subfigure}%
    \begin{subfigure}[b]{0.3\textwidth} 
        \centering
        \includegraphics[width=0.9\textwidth]{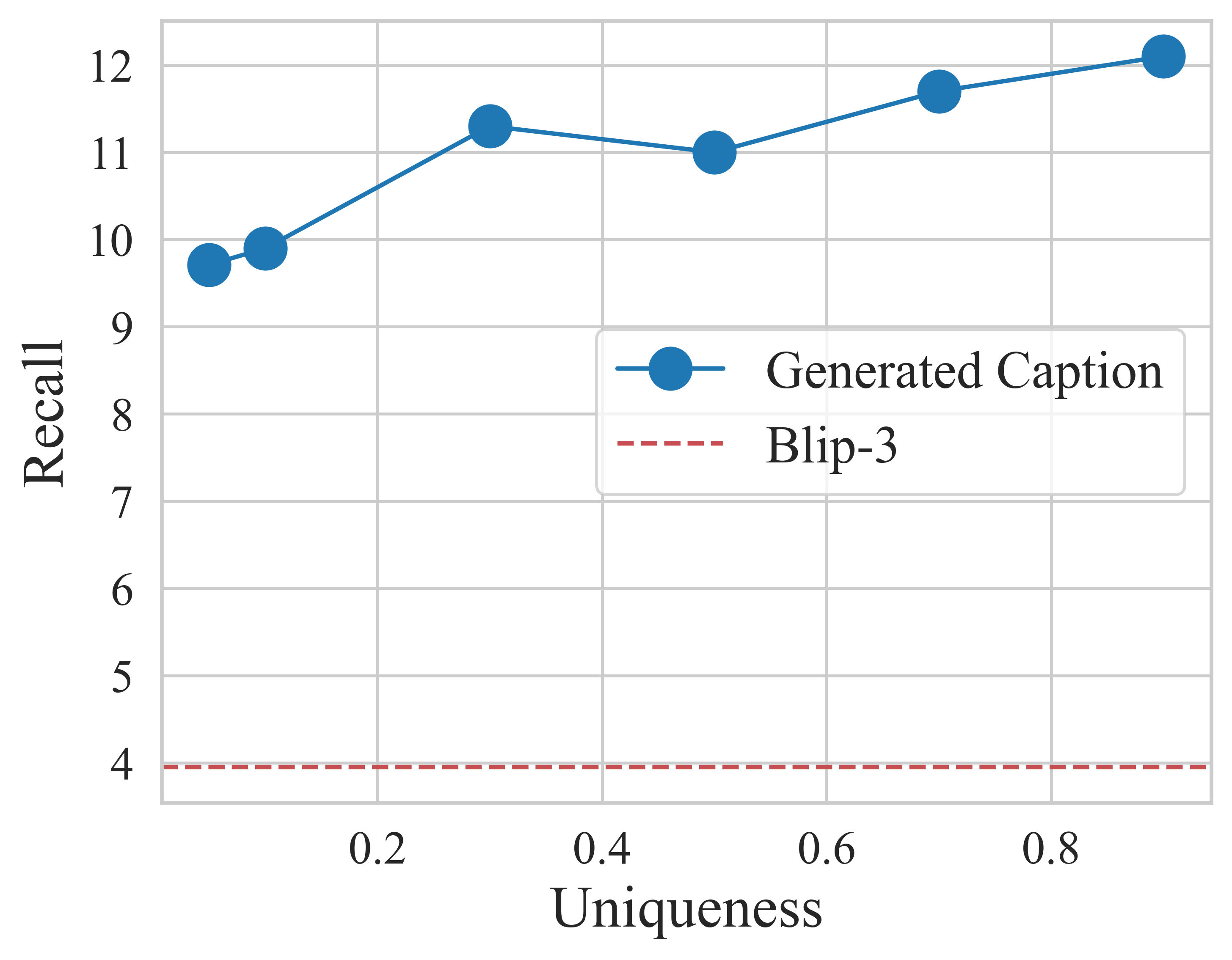} 
        \caption{Stanford Dog}
        \label{fig:sub2}
    \end{subfigure}
    \begin{subfigure}[b]{0.3\textwidth} 
        \centering
        \includegraphics[width=0.9\textwidth]{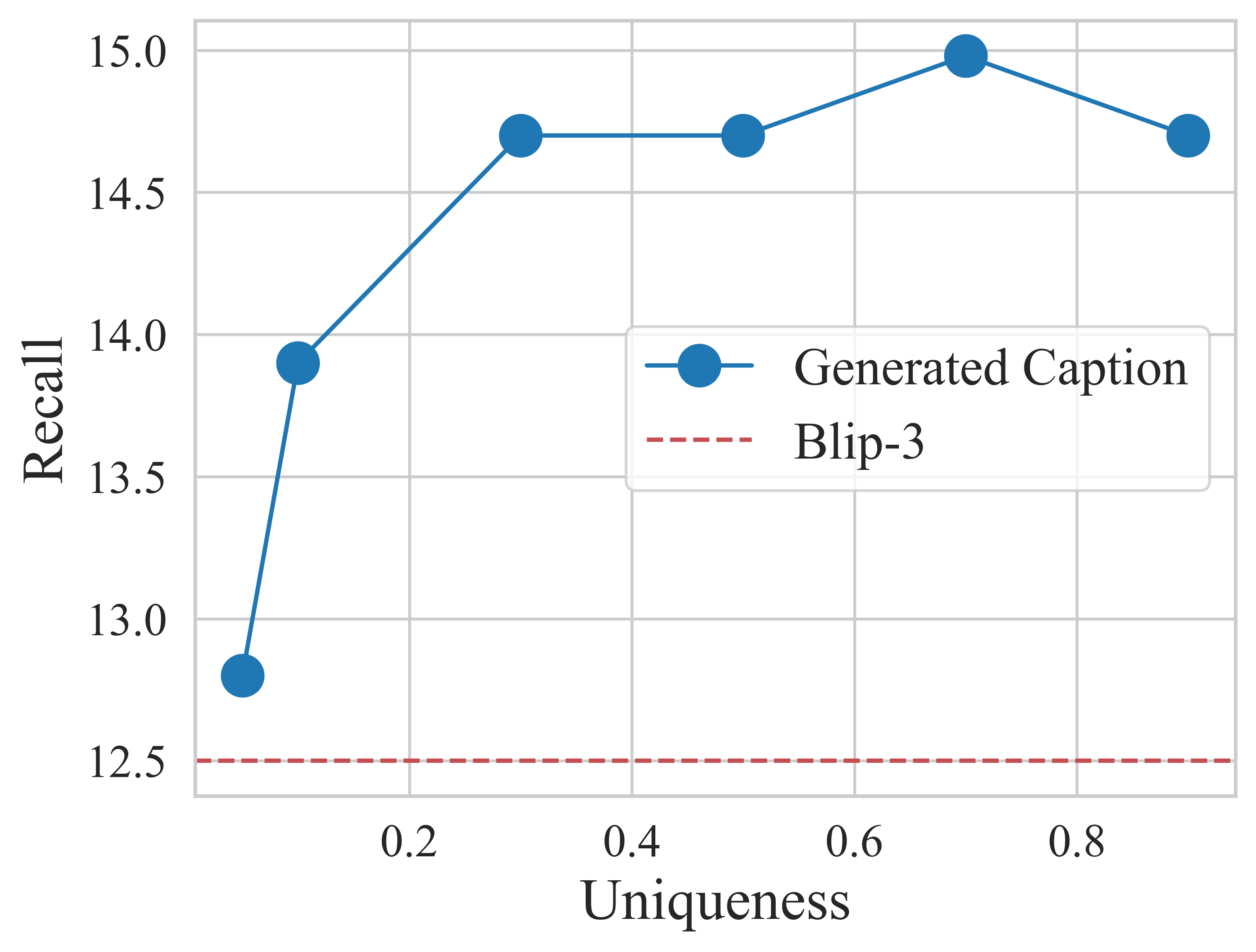} 
        \caption{Stanford Cars}
        \label{fig:sub3}
    \end{subfigure}
    \vspace{-3mm}
    \caption{The effect of varying uniqueness values. We assess if the output captions include fine-grained category names as recall, using three fine-grained category classification datasets, (a): CUB, (b): Stanford Dog, and (c): Stanford Cars. Increasing the uniqueness condition values improves recall, meaning that captions include fine-grained concept words as we intend.}
    \label{fig:uniqueness}
\end{figure*}

%% file: figs/example_uniqueness.tex
\begin{figure*}[htbp]
\vspace{-4mm}
    \centering
    \includegraphics[width=\linewidth]{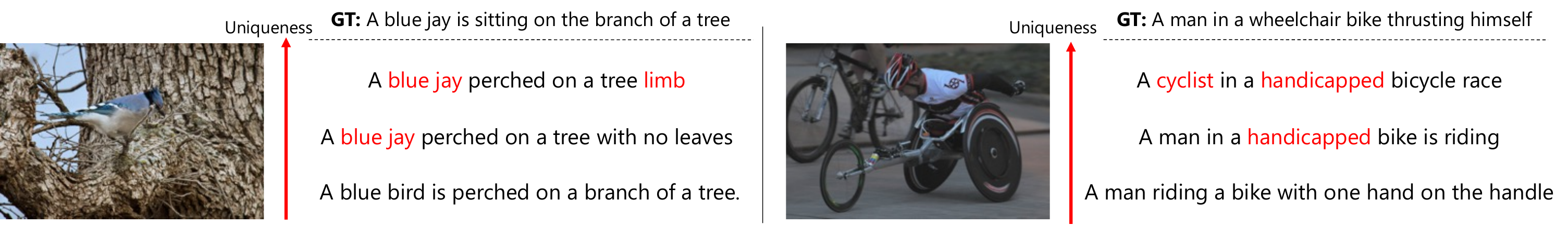}
        \vspace{-7mm}
    \caption{Generated captions by increasing a uniqueness conditioning value while fixing the other two conditions. Increasing the value can encourage the model to generate fine-grained category words. }
    \label{fig:example_uniqueness}
\end{figure*}

%% file: tables/self_retrieval.tex
\begin{table}[htbp]
    \begin{center}
    \scalebox{0.85}{
    \begin{tabular}{l|ccc|c}
        \toprule
        \multirow{1}{*}{Models} & R1 & R5 & R10 & Clipscore\\\midrule
GT Caption& 32.6&57.6&68.1&76.9\\\midrule
        BLIP-3~\cite{xue2024xgen} &25.9&49.8&60.3&73.9\\
        Vanilla-Supervised&30.9&56.5&67.5&77.6\\
        Concap~\cite{wang2023controllable}&32.5&58.4&69.4&77.9\\
        \ours &\textbf{36.9}&\textbf{62.2}&\textbf{72.7}&\textbf{78.9}\\       
        \bottomrule
    \end{tabular}
    }
    \end{center}
    \vspace{-0.55cm}
    \caption{
        \small Self-retrieval evaluation using COCO. 
    }
    \label{table:self_retrieval}
\end{table}

%% file: figs/vocabulary_bar.tex
\begin{figure}[t]
\vspace{-2mm}
    \centering
    \includegraphics[width=0.7\linewidth]{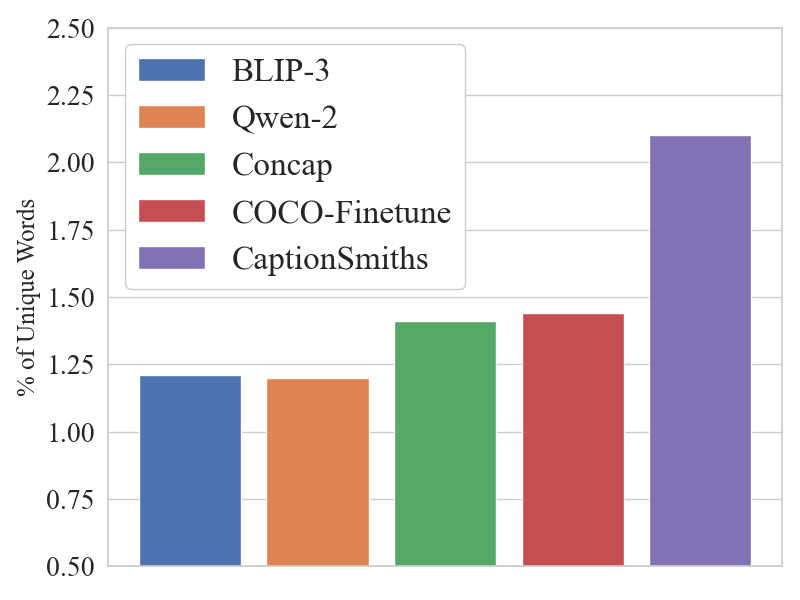}
    \vspace{-5mm}
    \caption{\small Comparison in the ratio of unique words in generated captions. The vocabularies of \ours are much richer than those of baselines.}
    \label{fig:vocabulary_var}
    \vspace{-2mm}
\end{figure}

%% file: figs/x_uniq_y_recall_vocab.tex
\begin{figure}[t]
    \centering
    \includegraphics[width=0.8\linewidth]{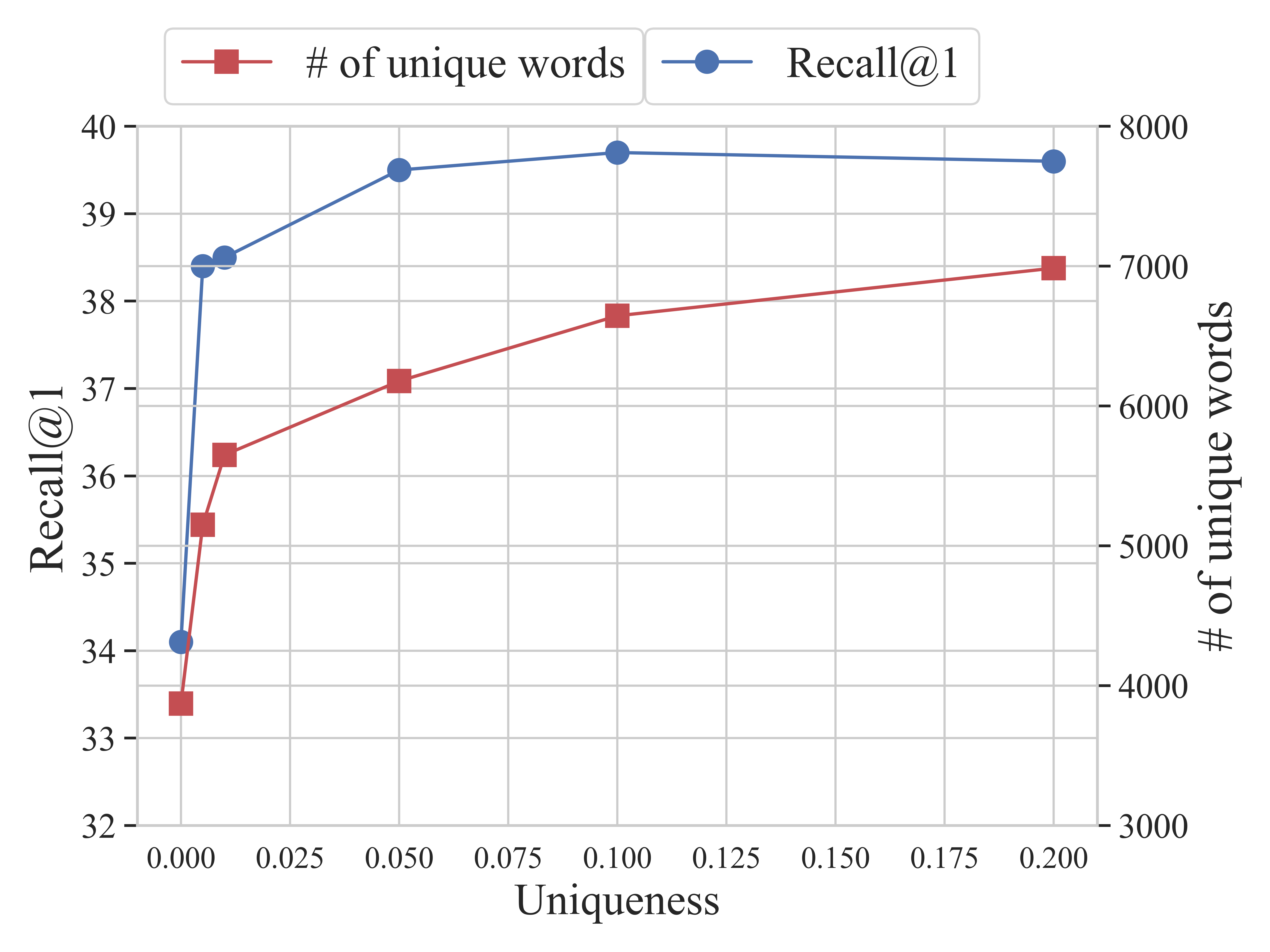}
    \vspace{-3mm}
    \caption{\small Increasing the uniqueness achieves richer vocabulary (right y-axis) and unique alignment with image (left y-axis).}
    \label{fig:x_uniq_y_recall_vocab}
\end{figure}

%% file: tables/coco_only.tex
\begin{table}[t]
\centering
\scalebox{0.85}{
\begin{tabular}{l|c|cccc}
\toprule
Models &Target-aware& B@4 &M &C &R \\
\midrule
Vanilla &&11.6&36.6&96.5&39.9\\
LC~\cite{deng2020length} &\checkmark& 11.8 & 37.1&103.6&40.3\\
\ours &\checkmark&\textbf{12.5}&\textbf{37.7}&\textbf{112.6}&\textbf{41.2}\\
\bottomrule
\end{tabular}}
\vspace{-3mm}
\caption{Comparison with conditioning baseline, LC~\cite{deng2020length}, in training with COCO dataset.}
\label{tab:coco_only}
\end{table}

%% file: tables/length_difference.tex
\begin{table}[t]
\centering
    \scalebox{0.85}{

\begin{tabular}{c|ccc|c}
\toprule
\textbf{Number of clusters} & \textbf{5} & \textbf{20} & \textbf{100} & \textbf{\ours} \\ \midrule
CIDEr ~$\uparrow$ & 110 & 105 & 103 & \textbf{112} \\ \hline
BLEU@4~$\uparrow$  & 12.5 & 11.9 & 11.6 & \textbf{12.6} \\ \hline
Rouge~$\uparrow$  & 40.8 & 40.2 & 39.8 & \textbf{41.2} \\ \hline
Meteor~$\uparrow$  & 37.8 & 37.2 & 36.0 & \textbf{37.8} \\ \hline
Length mismatch~$\downarrow$ & 13.8 & 10.9 & 9.7 & \textbf{1.6} \\ \bottomrule
\end{tabular}
}
\vspace{-3mm}
\caption{Comparison of metrics across different group sizes and our method. Length mismatch computes the difference between generated and ground-truth captions in their token counts. }
\label{tab:length_difference}
\end{table}

%% file: chapters/10_conclusion.tex
\vspace{-3mm}
\section{Conclusion}
We present an approach to build a single model capable of \textit{smoothly} adapting its behavior for three properties: length, descriptiveness, and uniqueness of words. \ours achieves slider-like flexible control over three properties. Empirical analysis shows that (i) \ours flexibly switches its state and outperforms baseline models when evaluated on diverse image-caption datasets. Although we focus on three properties to control the attribute of captions, controlling other aspects is interesting and our approach should be applicable. 

\noindent\textbf{Acknowledgement.} This work is supported by JST Moonshot R\&D Program Grant Number JPMJMS2236. Donghyun Kim was supported by the National Research Foundation of Korea(NRF) grant funded by the Korea government(MSIT)(No. RS-2024-00341514) and Institute of Information \& communications Technology Planning \& Evaluation (IITP) grant funded by the Korea government(MSIT) (No. RS-2019-II190079,  Artificial Intelligence Graduate School Program(Korea University)).

\label{sec:conclusion}

%% file: appendix_content.tex
\section{Addition Details on Method}
\noindent\textbf{Details on decorrelation.}
We describe the details on decorrelating properties using linear regression. 
To decorrelate the properties, we apply linear regression to model the dependency between properties and remove the correlation. First, we decorrelate the uniqueness (\(U_c\)) with respect to the caption length (\(L_c\)) by training a regression model. Let \(f_L(L_c)\) be the regression function that predicts \(U_c\) based on \(L_c\). We then subtract this predicted value from the original uniqueness:

\[
U_{c}^{\text{decorr}} = U_c - f_L(L_c)
\]

Next, we decorrelate the density (\(D_c\)) using both the length (\(L_c\)) and the processed uniqueness (\(U_c^{\text{decorr}}\)). The regression model \(f_{L,U}(L_c, U_c^{\text{decorr}})\) predicts \(D_c\) from both \(L_c\) and \(U_c^{\text{decorr}}\). We subtract this predicted value from the original density:

\[
D_{c}^{\text{decorr}} = D_c - f_{L,U}(L_c, U_c^{\text{decorr}})
\]

In these equations, \(f_L(L_c)\) and \(f_{L,U}(L_c, U_c^{\text{decorr}})\) are regression models trained to predict one property from the others. By removing these predicted components, we effectively reduce the correlation between the properties while preserving their interpretability. This ensures that variations in one property minimally affect the others, allowing independent control over each property.

\noindent\textbf{Excluded sets in descriptiveness calculation.} 
In computing the descriptiveness score, we exclude some nouns including “image", “side", “background", “picture", “top", and “bottom", since these nouns are included in many captions and are not very descriptive. 

\section{Experimental Details}
We will publish codes used for our experiments including the dataset split and trained weights upon acceptance.

\noindent\textbf{Training details.}
We employ the configurations used in LLaVA github repository\footnote{\url{https://github.com/haotian-liu/LLaVA}}.
The training is done in two stages; the first stage tunes the projector modules and condition embeddings, and the second stage also tunes both LM's parameters. We train models for one epoch in each stage since increasing the training epochs does not improve the performance. We abbreviate details on other hyper-parameters, \eg, batch size, design of the projector, and learning rate, since we follow the default hyper-parameters. All models are trained with 8 A100 GPUs with 80GB or 40GB memory. 

\noindent\textbf{Architecture details.}
For the visual encoder, we employ the openai's vit-large-patch14 model~\footnote{\url{https://huggingface.co/openai/clip-vit-large-patch14}}. Although llava-1.5 utilizes a larger model, vit-large-patch14-336~\footnote{\url{https://huggingface.co/openai/clip-vit-large-patch14-336}}, we employ the smaller one for the computational efficiency. 
Llama-2-7b-chat model\footnote{\url{https://huggingface.co/meta-llama/Llama-2-7b-chat-hf}} is used as a language decoder. 

\input{tables/dataset_length}

\noindent\textbf{Datasets.}
Table~\ref{tab:datasets} shows the number of image-caption pairs and average token number per dataset used in our experiments. We choose these datasets to cover captions with a wide range of length and vocabularies. For Localized Narrative and COCO evaluation, we employ the COCO's validation split for evaluation. For Docci, we employ its test split (2000 captions). 

\input{figs/vary_conditions_summary}

\noindent\textbf{LLaVA-1.5.} We employ the model in huggingface\footnote{\url{https://huggingface.co/liuhaotian/llava-v1.5-7b}}. To generate a caption, we switch concise/detailed prompts, \ie, “Please provide a short description.” and “Describe the image in detail.”. The former prompt is used for COCO and LNCOCO while the latter is used for Docci. However, we do not see the significant change in the caption style probably because LLaVA-1.5 is trained primarily on long captions. 

\noindent\textbf{BLIP-3.} We employ the model in huggingface\footnote{\url{https://huggingface.co/Salesforce/xgen-mm-phi3-mini-instruct-r-v1}}. To generate a caption, we switch concise/detailed prompts, \ie, “Please provide a short description.” and “Describe the image in detail.”. The former prompt is used for COCO and LNCOCO while the latter is used for Docci.

\noindent\textbf{Qwen2-VL.} We employ the model in huggingface\footnote{\url{https://huggingface.co/Qwen/Qwen2-VL-7B-Instruct}}. The same prompts as BLIP-3 are used for evaluation.

\noindent\textbf{Clip-score.} 
We employ openai/clip-vit-base-patch16 model to compute the score. This score computes the similarity between paired image and text, and scales by a constant value. 

\noindent\textbf{Self-retrieval.} We evaluate the performance to retrieve a paired image using a generated caption. Specifically, we employ the COCO validation set and assess if a caption can retrieve a paired image from all validation images, approximately 5000 in total.

\section{Additional Results}
\input{tables/nocaps}
\noindent\textbf{Zero-shot evaluation.} 
Table~\ref{table:zeroshot} shows the results on Nocaps~\cite{agrawal2019nocaps} and Vizwiz~\cite{gurari2020captioning}. The results suggest that ours is generalizable in diverse image captioning datasets. 

\begin{figure}
    \centering
    \includegraphics[width=\linewidth]{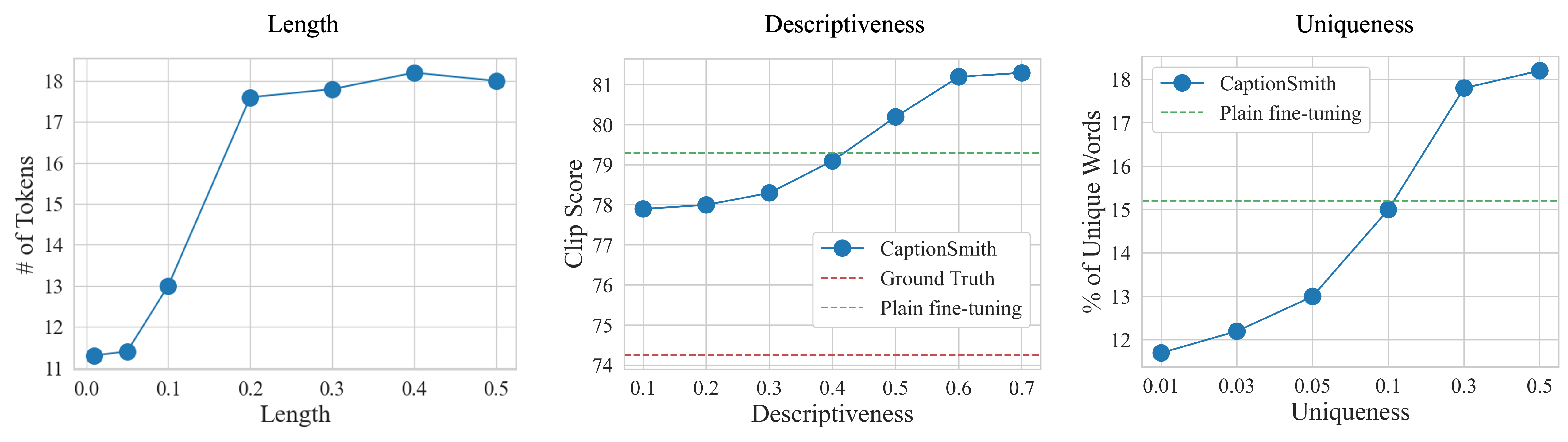}
    \vspace{-7mm}
    \caption{Results of fine-tuning Qwen-VL-2.5 32B with LoRA to manipulate length, descriptiveness, and uniqueness (from left to right).}
    \vspace{-5mm}
    \label{fig:qwen}
\end{figure}
\noindent\textbf{Generalization in different backbones.} 
We conduct fine-tuning Qwen-VL-2.5-32B\footnote{https://huggingface.co/Qwen/Qwen2.5-VL-32B-Instruct} for CaptionSmiths. Due to the limitations of time and computation, we train the model on COCO with LoRA tuning.  Fig.~\ref{fig:qwen} demonstrates that the trained model can manipulate three attributes as the LLaVA backbone does in the main draft. This is strong empirical evidence that our framework is applicable to diverse models, including large ones.

\noindent\textbf{Three conditions are well-disentangled.} 
In our framework, we expect the model not to change the other two properties when changing one condition and fixing others. Fig.~\ref{fig:vary_conditions_summary} studies how varying one condition affects other properties. In summary, varying one property slightly affects others, but the effect is insignificant. For instance, the left of Fig.~\ref{fig:vary_conditions_summary} shows the change in the length of output captions in varying uniqueness conditions. The change lies within 1-2 tokens, which is insignificant. In the other two cases, where we visualize the range of properties computed in COCO validation captions as a reference, the change is also insignificant. Although not perfect, our conditioning achieves well-disentangled control in three properties.

\input{figs/ablation_decorrelation}
\noindent\textbf{Ablation study for decorrelation.} 
We conduct an ablation study on the decorrelation process to compute conditioning values in Fig.~\ref{fig:ablate_uniqueness}. We vary the uniqueness value as done in Fig. 8 of the main draft. In both cases, increasing uniqueness tends to increase the recall. However, the improvement is not very clear in \textit{Not decorrelated} approach evaluated on Stanford Dog. Additionally, applying decorrelation improves overall recall. Since three properties used in our experiments are not highly correlated, directly using computed values might suffice. However, when adding more conditioning, the correlation between properties can be a concern and we provide a potential solution to it.

\input{tables/ablation_three}
\noindent\textbf{Effect of three properties on lexical alignment.} 
We conduct an ablation study on the effect of controlling three properties for lexical alignment in COCO. Specifically, we employ a model in Table 1 and ablate controlled properties. For instance, to ablate the control of descriptiveness, we use the average value of the descriptiveness computed in COCO for all samples. Table~\ref{tab:ablation_properties} shows that adding control consistently improves the lexical alignment. 

\input{figs/long_cap_example}
\noindent\textbf{Hallucination in long-captions.}
Fig.~\ref{fig:long_caption} shows the example of an input image and generated caption. The model hallucinates about the tail of the cat though most details are correct. As shown in previous studies~\cite{hirota2024descriptive}, generating a long caption without hallucination is challenging, and the problem is not addressed by our approach. We highlight that addressing this issue is not the main scope of our work and leave it for our future work. 

\input{figs/continuous_discrete}

\noindent\textbf{Continuous vs. discrete conditioning.} 
Fig.~\ref{fig:discrete_vs_continuous} shows the alignment with ground-truth captions in terms of the length. Captions generated by our approach clearly show better alignment than the ones generated by discrete conditioning. Since discrete conditioning groups captions with different lengths into the same mode, the model cannot capture their unique length during generation. 

\input{figs/descriptive_gptscore}

\noindent\textbf{Evaluation with GPT.} 
Our conditioning is obtained with our-defined ways of computation. However, the criterion is not necessarily aligned with the human criterion. Since conducting human study involves a lot of efforts and reproducibility issues, we employ ChatGPT for this purpose, motivated by the fact that ChatGPT and human judgements of a caption are well-aligned~\cite{chan2023clair}.
Following Chan\etal~\cite{chan2023clair}, we prompt ChatGPT to return the score ranging from 0 to 100 as shown in Fig.~\ref{fig:prompt_chatgpt}.
\captionsetup{hypcap=false}
\begin{center}
\begin{minipage}{0.9\linewidth} 
\begin{verbatim}
You will be given a caption describing an image.
Your task is judging the detailness of the captions.
If you think the caption is detailed, score will get higher.
Your score needs to be in range of 0 to 100.
Return your score only, e.g., 80.
Caption: {caption}
Your answer:
\end{verbatim}
\captionof{figure}{Prompt sentence used to evaluate generated caption's descriptiveness.}
\label{fig:prompt_chatgpt}
\end{minipage}
\end{center}
Fig.~\ref{fig:gpt_descriptive} illustrates the results with different descriptiveness conditions. We observe that ChatGPT gives a higher score to captions generated with a higher descriptiveness condition value. This demonstrates that our descriptiveness criterion matches with that of ChatGPT.

\noindent\textbf{More examples in varying descriptiveness.} 
Fig.~\ref{fig:example_descriptive_appendix} illustrates results on varying descriptiveness conditioning. Increasing the descriptiveness score enriches the content in the caption. 

\input{figs/example_descriptive_appendix}

\input{figs/example_docci_appendix}

\noindent\textbf{Examples in captioning with many sentences.} 
Fig.~\ref{fig:example_docci_appendix} presents the example of Docci~\cite{onoe2025docci} ground-truth captions and captions generated by our approach. The first example does not show clear hallucinations while the model suffers from hallucinations in the second one. We observe that the model tends to hallucinate on the small details of the image.

%% file: tables/dataset_length.tex
\begin{table}[h!]
\centering
\begin{tabular}{l|cc}
\toprule
\textbf{Dataset} & \textbf{\# of Samples} & \textbf{\# 
 of Tokens} \\
\midrule
Localized Narrative~\cite{pont2020connecting} & 690K &46  \\
Detail23K~\cite{liu2024visual} & 23K & 140\\
Docci~\cite{onoe2025docci} & 10K &155\\
Laion-COCO~\cite{laion_coco} & 270K & 13 \\
COCO~\cite{lin2014microsoft} & 100K &  15\\
Monkey~\cite{li2024monkey} & 210K & 107\\
\bottomrule
\end{tabular}
\caption{Summary of datasets and their properties.}
\label{tab:datasets}
\end{table}

%% file: figs/vary_conditions_summary.tex
\begin{figure*}[htbp]
    \centering
    \includegraphics[width=\linewidth]{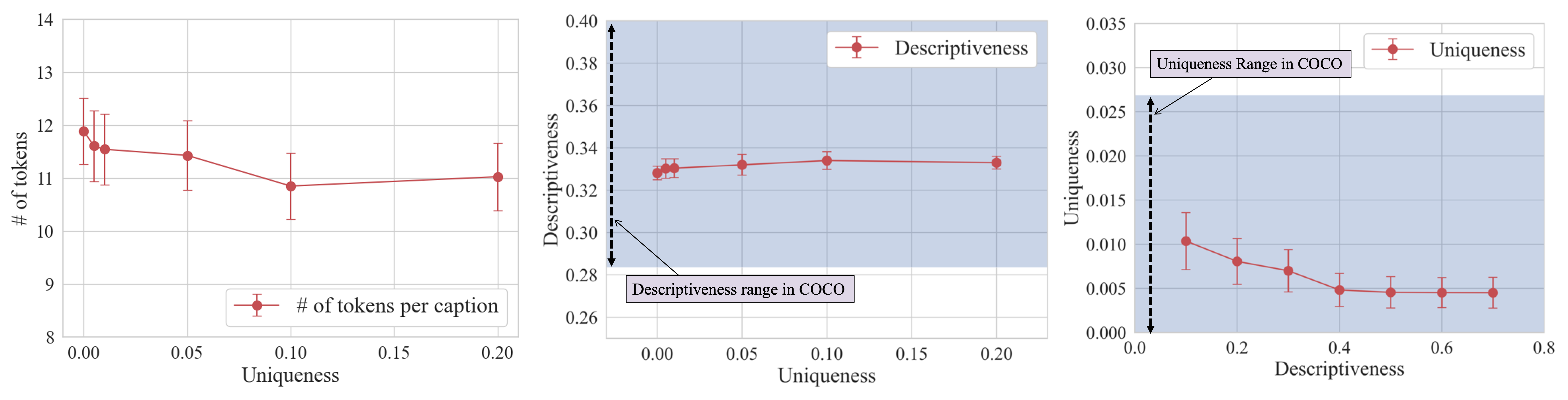}
    \vspace{-5mm}
    \caption{\small Study on varying one condition while fixing the other two conditions. \textbf{Left:} The change in the length by increasing uniqueness. Increasing uniqueness slightly shortens the number of tokens in generated captions. 
    \textbf{Middle:} The change in descriptiveness by increasing uniqueness. Considering the descriptiveness range visualized by the average and standard deviation in COCO GT captions (black dashed array), the amount of the change is not significant. \textbf{Right:} The change in uniqueness by increasing descriptiveness. The change is insignificant compared to the range in COCO GT captions (black dashed array).}
    \label{fig:vary_conditions_summary}
\end{figure*}

%% file: tables/nocaps.tex
\begin{table}[htbp]
    \begin{center}
    \scalebox{0.85}{
    \begin{tabular}{l|cccc|cccc|cccc|cccc}
        \toprule
        \multirow{3}{*}{Models} & \multicolumn{12}{c|}{No caps} & \multicolumn{4}{c}{Vizwiz } \\\cline{2-17}
        & \multicolumn{4}{c|}{in} & \multicolumn{4}{c|}{near} &  \multicolumn{4}{c|}{out} &\multicolumn{4}{c}{val} \\
        &B@4 &M &C &R  &B@4 &M &C &R &B@4 &M &C &R &B@4 &M &C &R \\\hline
        BLIP-3~\cite{xue2024xgen} & 7.3&26.4&51.9&33.5&6.5&26.9&53.1&\textbf{34.0}&\textbf{5.4}&24.6&54.4&\textbf{31.8}&3.3&19.0&32.4&\textbf{25.6}\\
        Qwen-2-VL~\cite{Qwen2-VL} & 2.7&30.7&27.5&25.0&2.8&30.7&29.9&25.1&2.2&29.5&29.4&23.9&2.3&\textbf{24.1}&23.9&23.1\\
        Ours & \textbf{7.9}&\textbf{34.1}&\textbf{75.0}&\textbf{34.2}&\textbf{7.6}&\textbf{33.8}&\textbf{75.0}&\textbf{34.0}&\textbf{5.4}&\textbf{30.4}&\textbf{63.9}&30.2&\textbf{3.5}&22.3&\textbf{43.1}&23.0\\
        \bottomrule
    \end{tabular}
    }
    \end{center}
    \vspace{-0.55cm}
    \caption{
        \small Zero-shot caption generation evaluation with NoCaps~\cite{agrawal2019nocaps} and Vizwiz~\cite{gurari2020captioning}. We employ CIDEr, Rouge, Bleu-4, and Meteor as evaluation metrics for `in`, `near`, `out`, and `val`.
    }
    \label{table:zeroshot}
\end{table}

%% file: figs/ablation_decorrelation.tex
\begin{figure*}[htbp]
    \centering
    \begin{subfigure}[b]{0.45\textwidth} 
        \centering
        \includegraphics[width=0.8\textwidth]{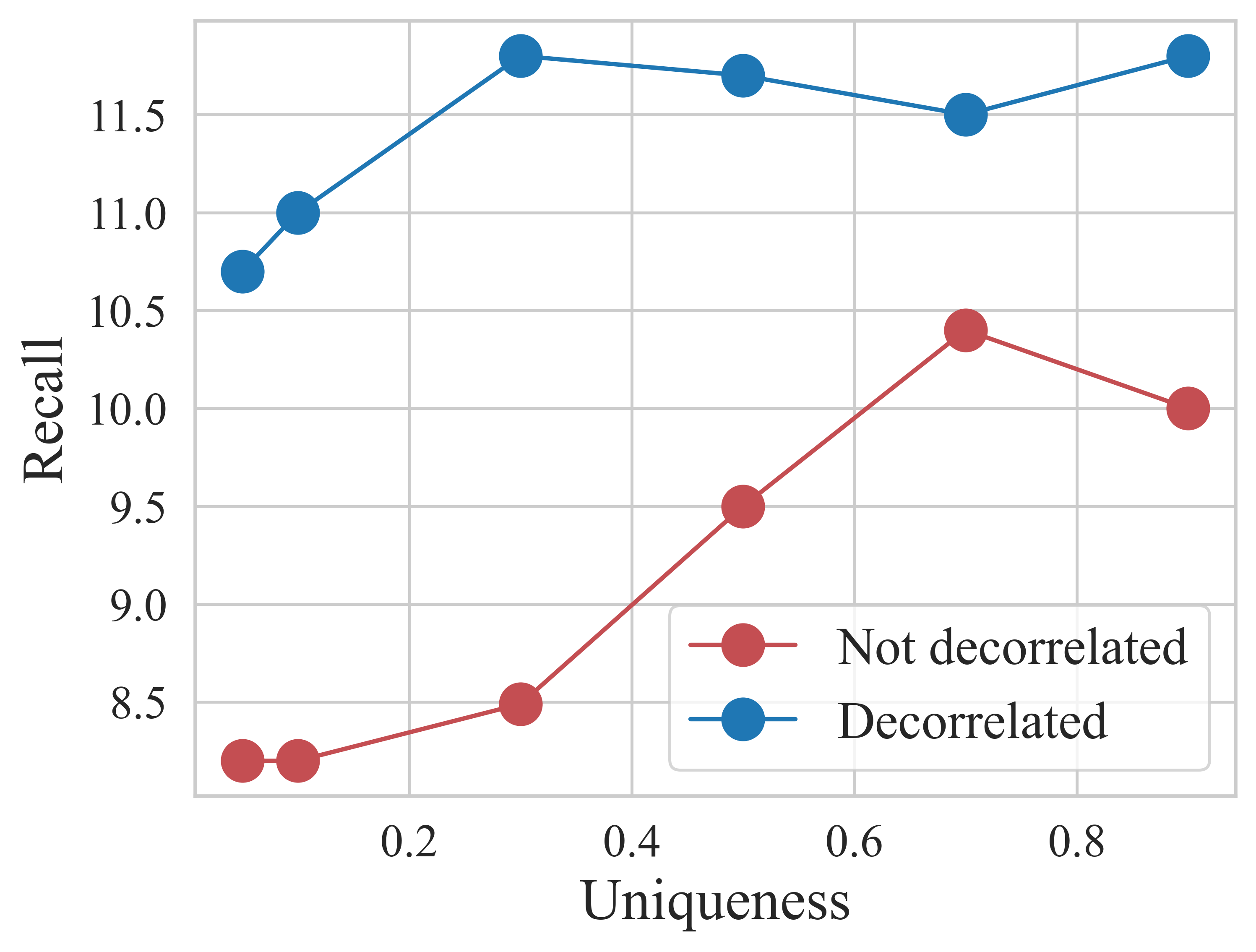} 
        \caption{CUB}
    \end{subfigure}%
    \begin{subfigure}[b]{0.45\textwidth} 
        \centering
        \includegraphics[width=0.8\textwidth]{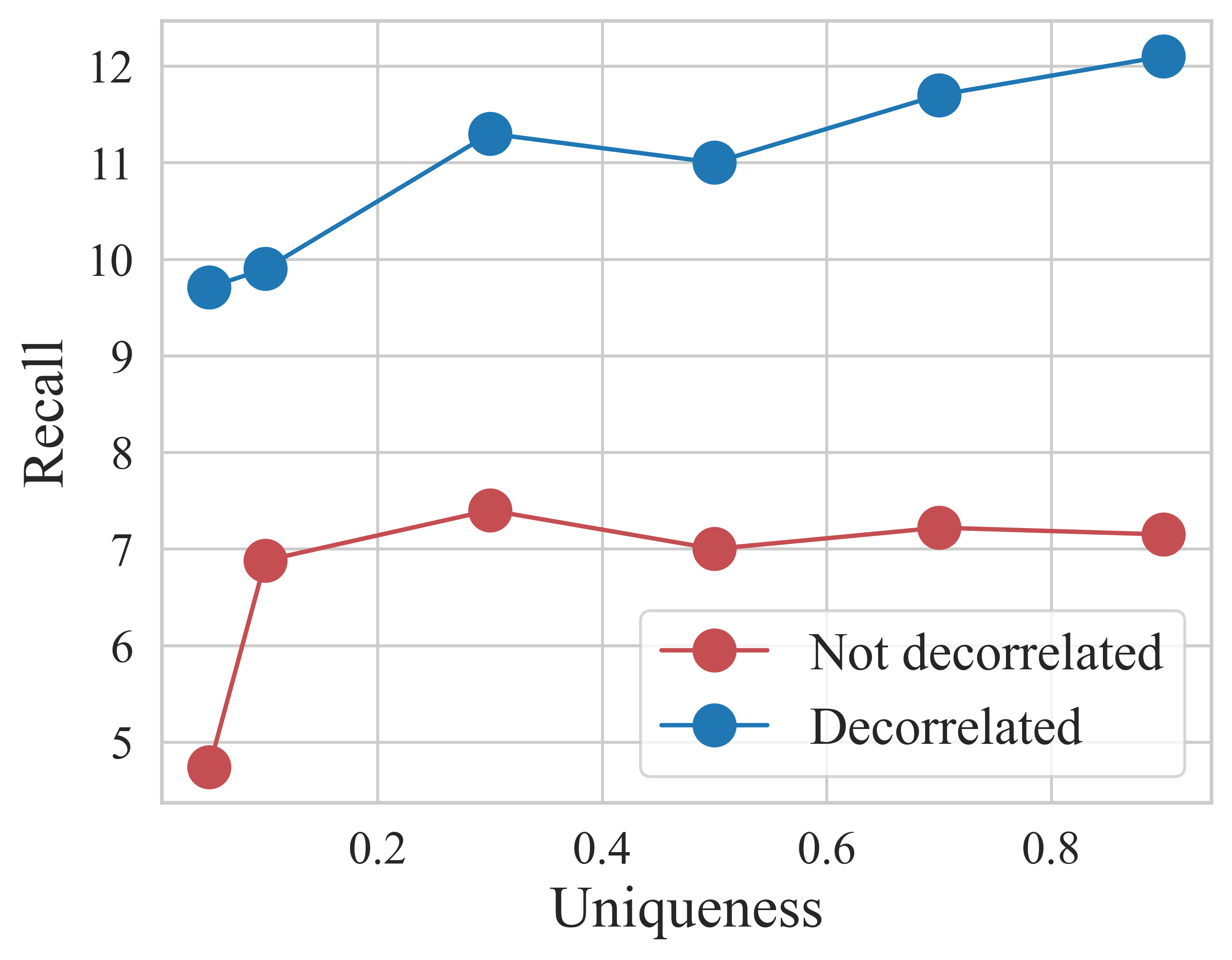} 
        \caption{Stanford Dog}
    \end{subfigure}
    \vspace{-3mm}
    \caption{Ablation study on decorrelation.}
    \label{fig:ablate_uniqueness}
\end{figure*}

%% file: tables/ablation_three.tex
\begin{table}[h!]
\centering
\scalebox{1.0}{
\begin{tabular}{ccc|cccc}
\toprule
Length&Descriptiveness&Uniqueness & B@4 &M &C &R \\
\midrule
\checkmark&&&9.7&36.2&89.0&38.0\\
\checkmark&\checkmark&&10.8&37.8&98.8&39.3\\
\checkmark&&\checkmark&10.2&37.1&95.1&38.6\\
\checkmark&\checkmark&\checkmark&\textbf{11.4}&\textbf{38.8}&\textbf{104.8}&\textbf{39.8}\\
\bottomrule
\end{tabular}}
\caption{The effect of controlling three properties on lexical alignment.}
\label{tab:ablation_properties}
\end{table}

%% file: figs/long_cap_example.tex
\begin{figure}[htbp]
    \centering
    \includegraphics[width=0.8\linewidth]{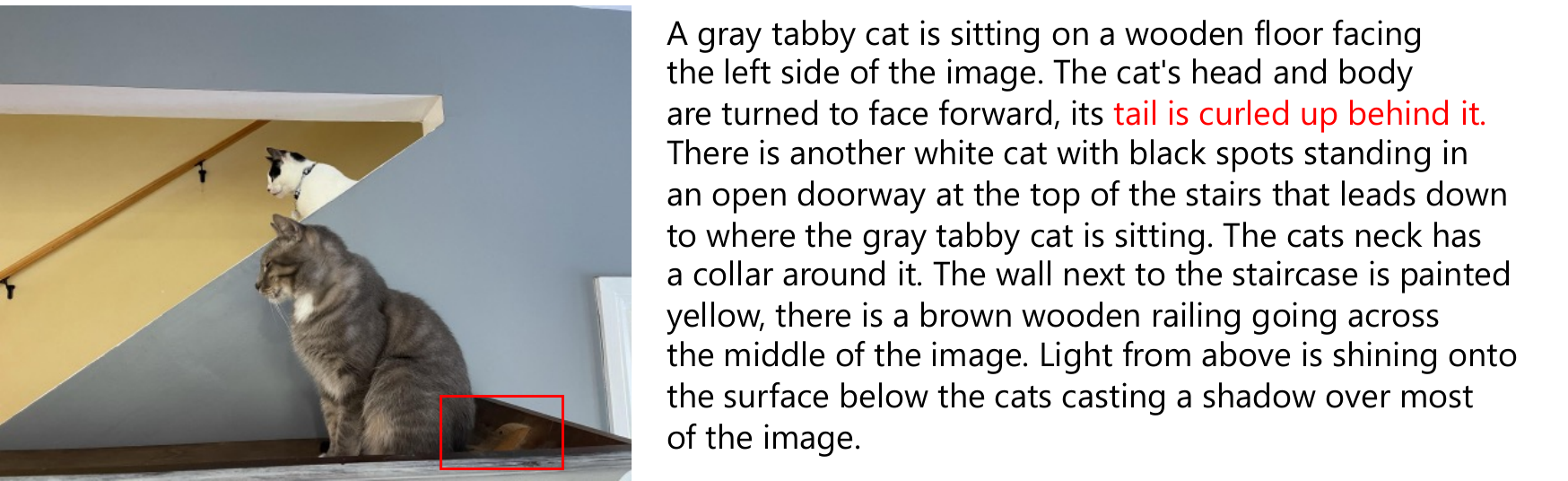}
    \caption{Hallunination long-caption. The model misunderstands that the brow leaf highlighted in the red box is the tail of the cat.}
    \label{fig:long_caption}
\end{figure}

%% file: figs/continuous_discrete.tex
\begin{figure}[t]
    \centering
    \includegraphics[width=0.9\linewidth]{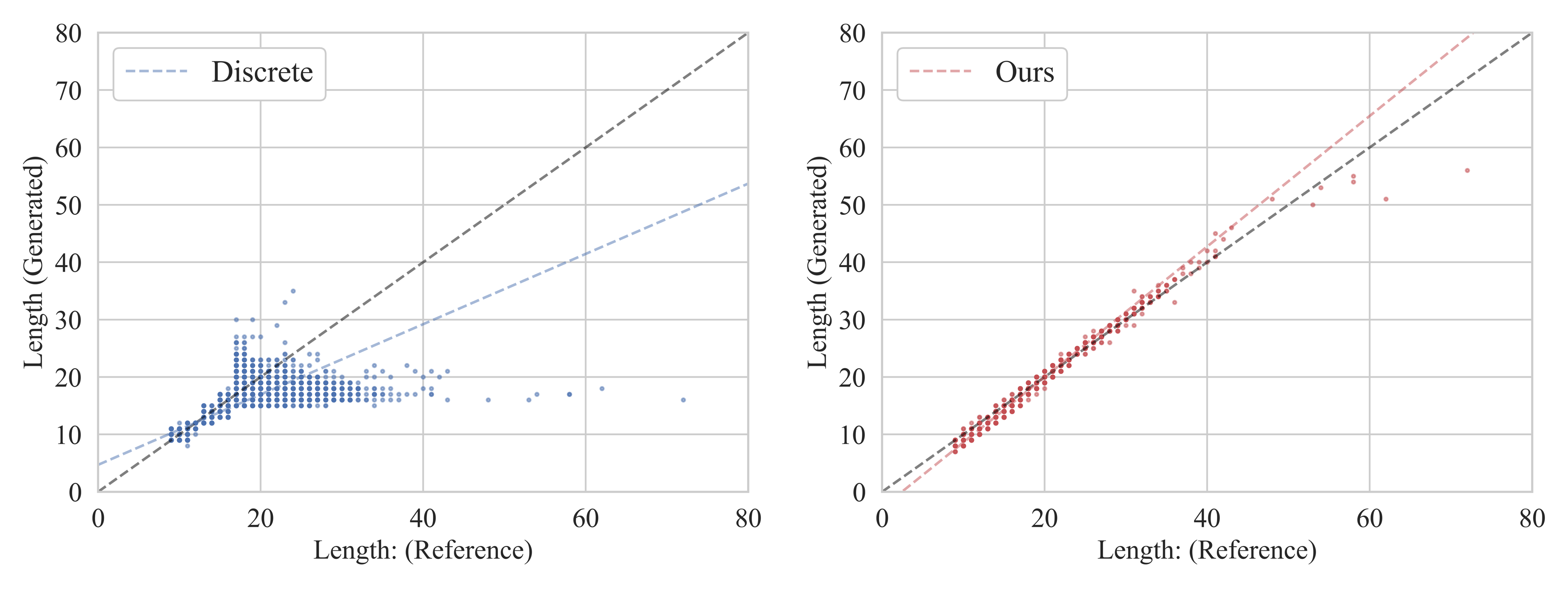}
    \caption{Comparison between discrete (\textbf{left}) and continuous (\textbf{right}) parameterization for conditioning. We find that continuous parameterization shows better alignment with respect to length. }
    \label{fig:discrete_vs_continuous}
\end{figure}


%% file: figs/descriptive_gptscore.tex
\begin{figure}[htbp]
    \centering
    \includegraphics[width=0.5\linewidth]{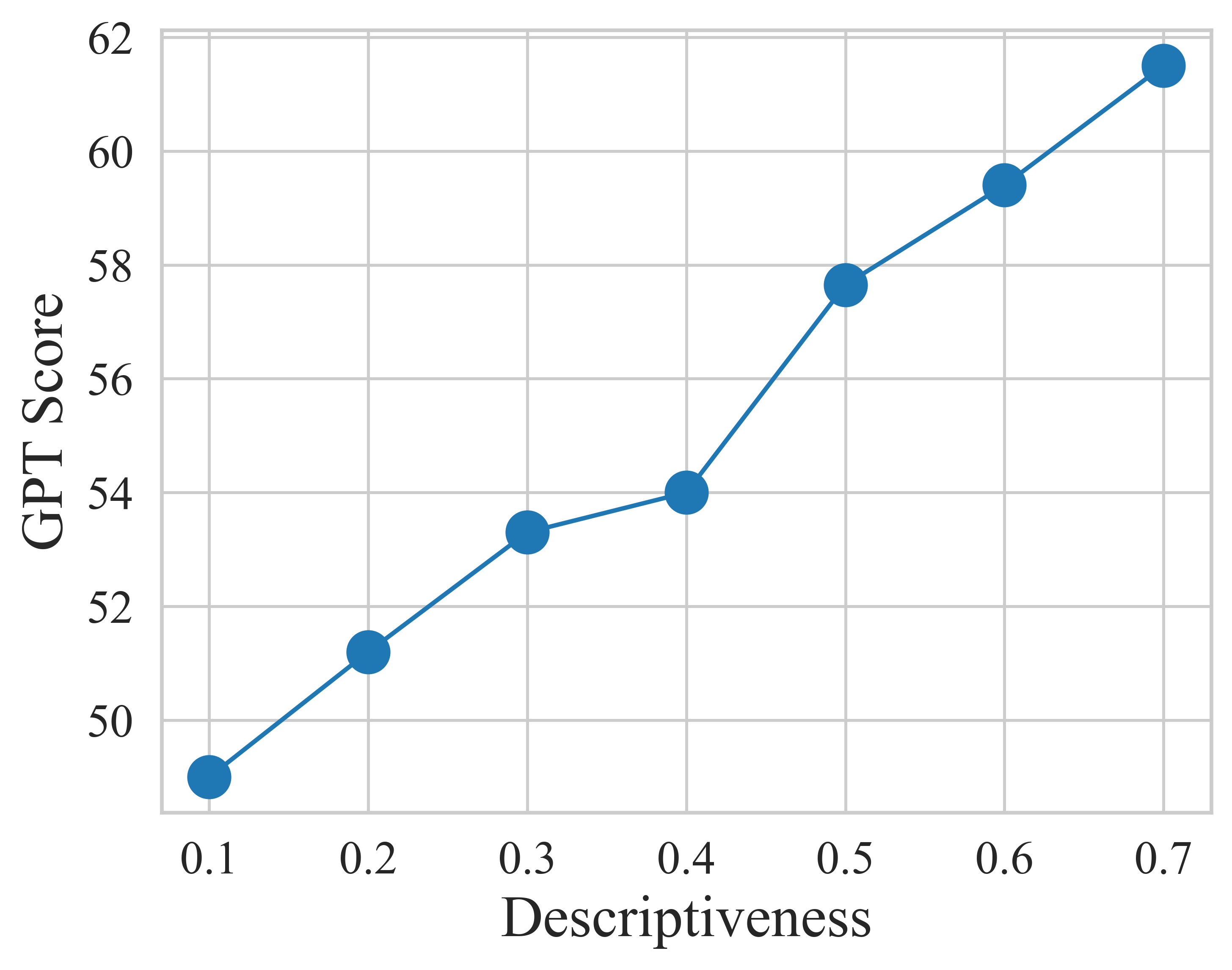}
    \caption{ChatGPT considers captions conditioned with larger descriptiveness to be ones with richer information.}
    \label{fig:gpt_descriptive}
\end{figure}

%% file: figs/example_descriptive_appendix.tex
\begin{figure*}[htbp]
    \centering
    \includegraphics[width=0.9\linewidth]{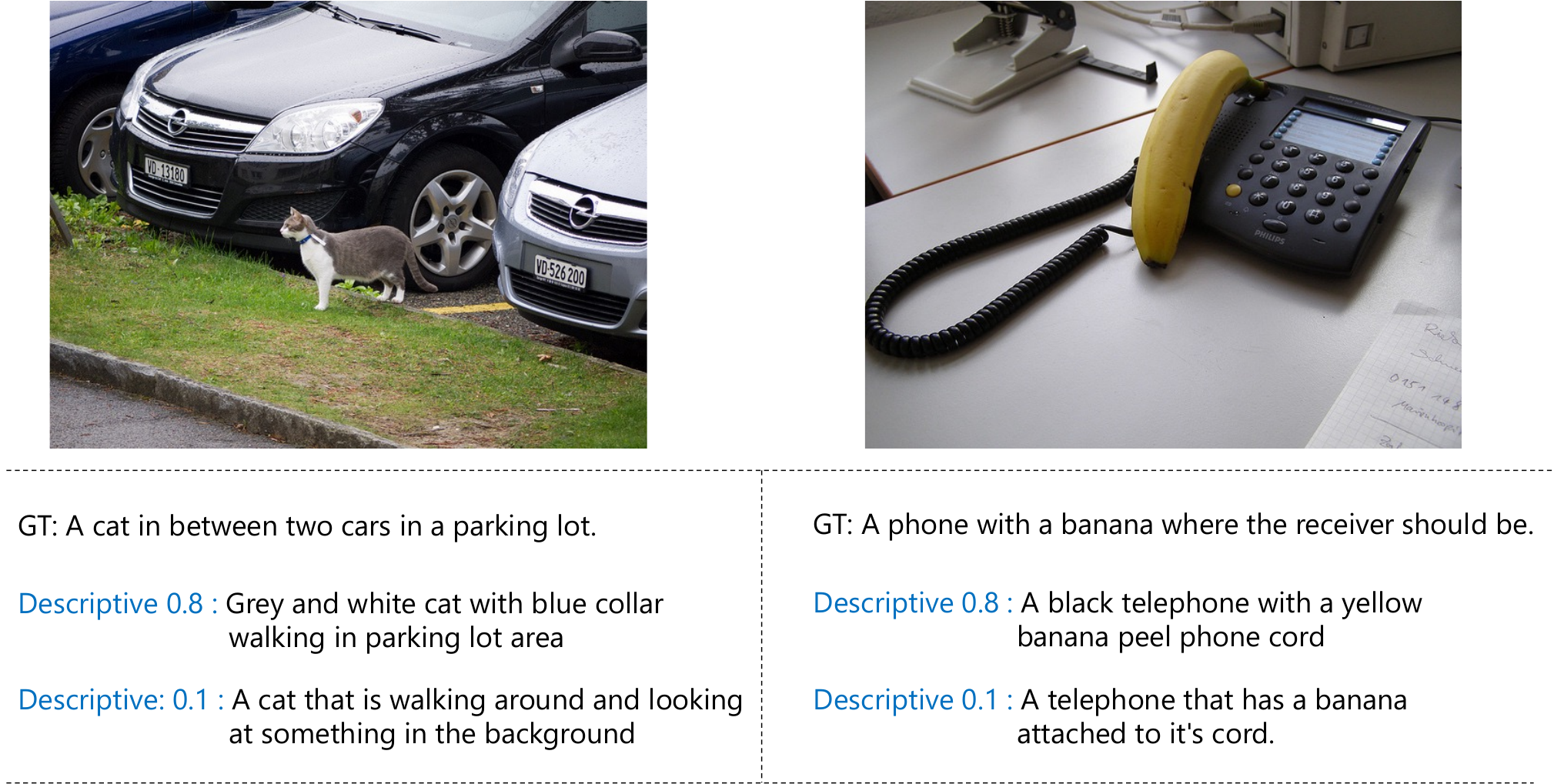}
    \caption{Generated captions by increasing a uniqueness conditioning value while fixing the other two conditions. Increasing the value can encourage the model to generate fine-grained category words. }
    \label{fig:example_descriptive_appendix}
\end{figure*}

%% file: figs/example_docci_appendix.tex
\begin{figure*}[htbp]
    \centering
    \includegraphics[width=0.8\linewidth]{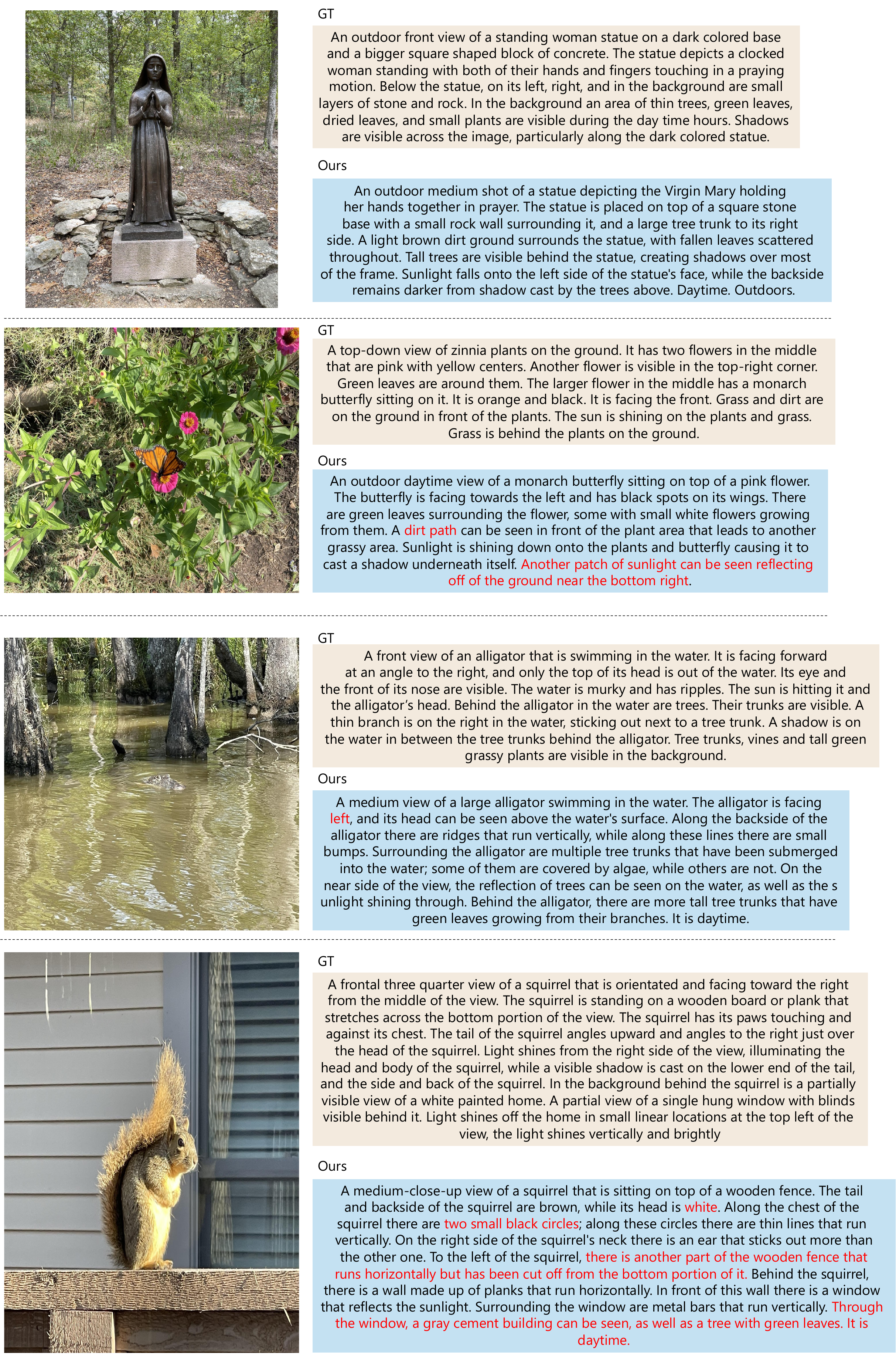}
    \caption{Generated captions for Docci~\cite{onoe2025docci}. Clear hallucinations are highlighted with red. }
    \label{fig:example_docci_appendix}
\end{figure*}